\documentclass[conference]{IEEEtran}
\usepackage{times}

\usepackage[numbers]{natbib}
\usepackage{multicol}
\usepackage{multirow}
\usepackage{amsfonts}
\usepackage{graphicx}
\usepackage{subfig}
\usepackage{xcolor}
\usepackage{hyperref}
\hypersetup{bookmarks=true}
\usepackage{duckuments}
\usepackage{booktabs}
\usepackage{float}
\usepackage{amsmath}
\let\labelindent\relax
\usepackage{enumitem}
 \usepackage[verbose]{placeins}
\usepackage[ruled,vlined,noend,linesnumbered]{algorithm2e}
\usepackage{makecell}
\definecolor{dark_yellow}{RGB}{252,194,3}

\begin{document}

% paper title
\title{Few-shot Adaptation for Manipulating Granular Materials Under Domain Shift}

\author{\authorblockN{Yifan Zhu$^*$, Pranay Thangeda$^*$, Melkior Ornik, Kris Hauser}
\authorblockA{University of
Illinois Urbana-Champaign\\
\texttt{\{yifan16,pranayt2,mornik,kkhauser\}@illinois.edu }\\
$^*$These authors contributed equally to this work }
}

\maketitle

\begin{abstract}
Autonomous lander missions on extraterrestrial bodies will need to sample granular material while coping with domain shift, no matter how well a sampling strategy is tuned on Earth. This paper proposes an adaptive scooping strategy that uses deep Gaussian process method trained with meta-learning to learn on-line from very limited experience on the target terrains. It introduces a novel meta-training approach, Deep Meta-Learning with Controlled Deployment Gaps (CoDeGa), that explicitly trains the deep kernel to predict scooping volume robustly under large domain shifts. Employed in a Bayesian Optimization sequential decision-making framework, the proposed method allows the robot to use vision and very little on-line experience to achieve high-quality scooping actions on out-of-distribution terrains, significantly outperforming non-adaptive methods proposed in the excavation literature as well as other state-of-the-art meta-learning methods. Moreover, a dataset of 6,700 executed scoops collected on a diverse set of materials, terrain topography, and compositions is made available for future research in granular material manipulation and meta-learning.

\end{abstract}

\IEEEpeerreviewmaketitle

\section{Introduction}

Terrain sampling is a key component of scientific exploration of planets and other extraterrestrial bodies~\cite{NASA2017}. However, Earth-bound teleoperation, as typically done in existing landers, faces intermittent and delayed communication that incurs latency of minutes or even hours in the case of long-duration interference. Autonomous sampling, in which the robot interprets sensor signals and makes decisions on where and how to sample, could drastically increase the efficiency of exploration. However, realizing autonomous sampling is challenging due to uncertainty in terrain material properties, composition, appearance, and geometry, limits in onboard computation, and a limited sampling budget.

\begin{figure}[h!]
\centering
\setlength{\tabcolsep}{0px}
\begin{tabular}{cc}
{\includegraphics[trim=0 0.41cm 0 0,clip,width=.4\linewidth]{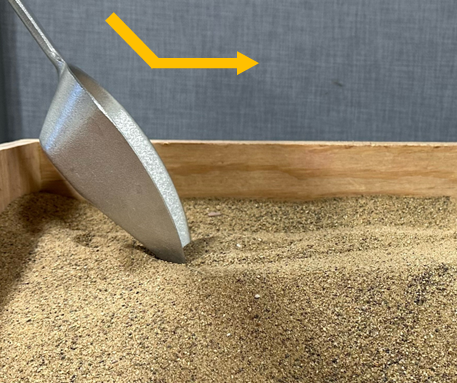}}\put(-115,65){\scalebox{0.8}{(a)}} &
{\includegraphics[width=.4\linewidth]{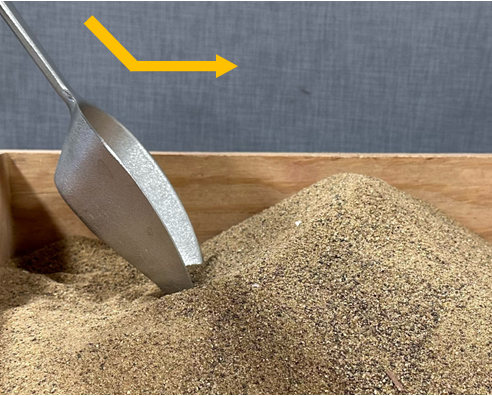}} \\
{\includegraphics[width=.4\linewidth]{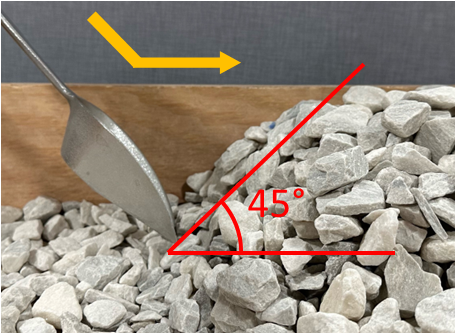}}\put(-115,67){\scalebox{0.8}{(b)}} &
{\includegraphics[trim=0 0.1cm 0 0,clip,width=.4\linewidth]{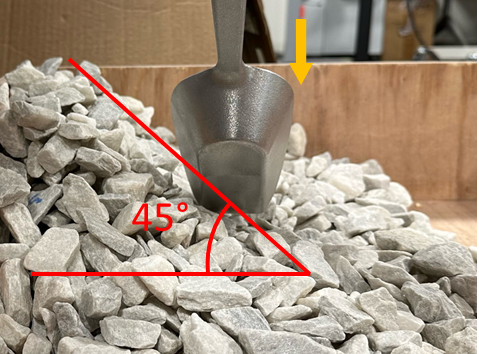}} \\
\multicolumn{2}{c}{\includegraphics[trim=14.5cm 9cm 13.5cm 10.5cm,clip,width=.8\linewidth]{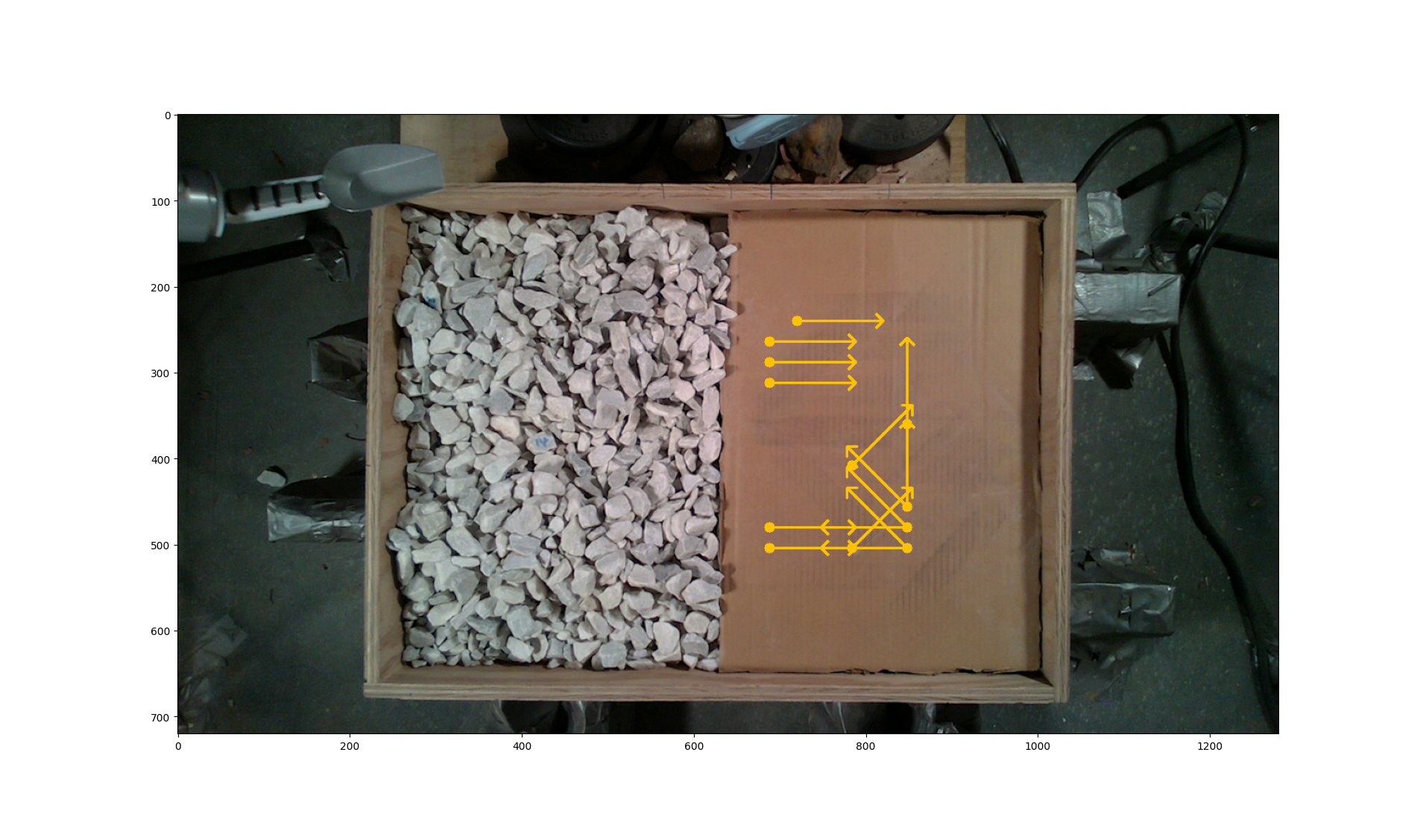}}\put(-215,80){\scalebox{0.8}{(c)}} \\
\multicolumn{2}{c}{\includegraphics[trim=14.5cm 11cm 13.5cm 9.5cm,clip,width=.8\linewidth]{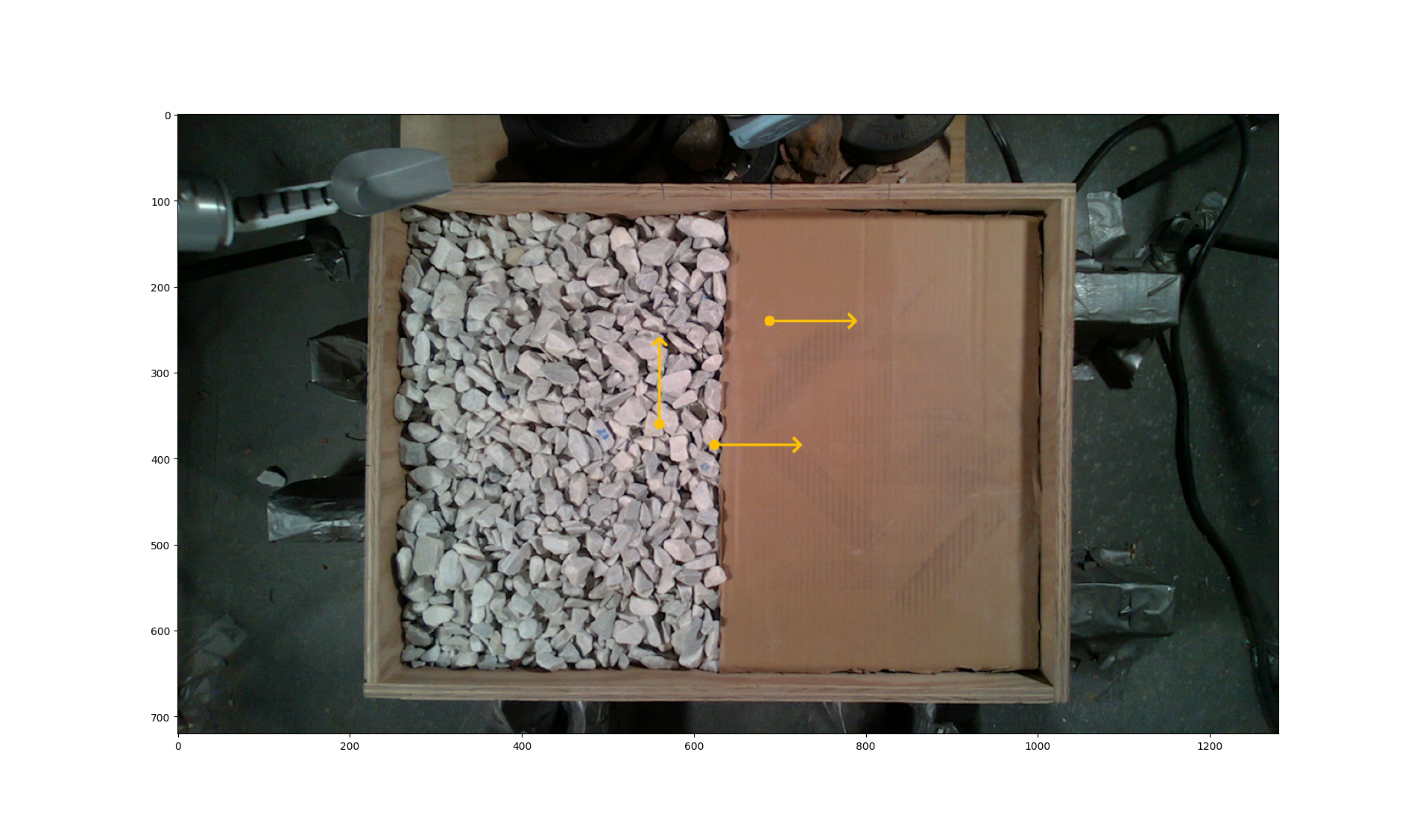}}\put(-215,70){\scalebox{0.8}{(d)}} \put(-80,70){\scalebox{1}{\textcolor{dark_yellow}{Scoop 1}}} \put(-100,30){\scalebox{1}{\textcolor{dark_yellow}{Scoop 2}}}
\put(-115,50){\scalebox{1}{\textcolor{dark_yellow}{Scoop 3}}}

\end{tabular}
\caption{(a) Scooping sand towards a slope results in a larger scooped volume than scooping a flat surface, but (b) scooping gravel towards a slope (left) often results in jamming due to interlocking between particles. Instead, scooping perpendicularly to a slope (right) causes a large volume of rocks to fall into the scoop without excessive effort. (c) On a terrain with a novel non-scoopable material (right), a non-adaptive strategy predicts high volumes on this out-of-distribution material and only selects to scoop there (yellow arrows). (d) Our adaptive strategy adjusts based on the feedback to explore novel and uncertain regions [Best viewed in color].}\label{fig:scooping_strategy}
\vspace{-0.5cm}
\end{figure}

\begin{figure*}[h!]
    \centering
     \includegraphics[width=0.9\linewidth]{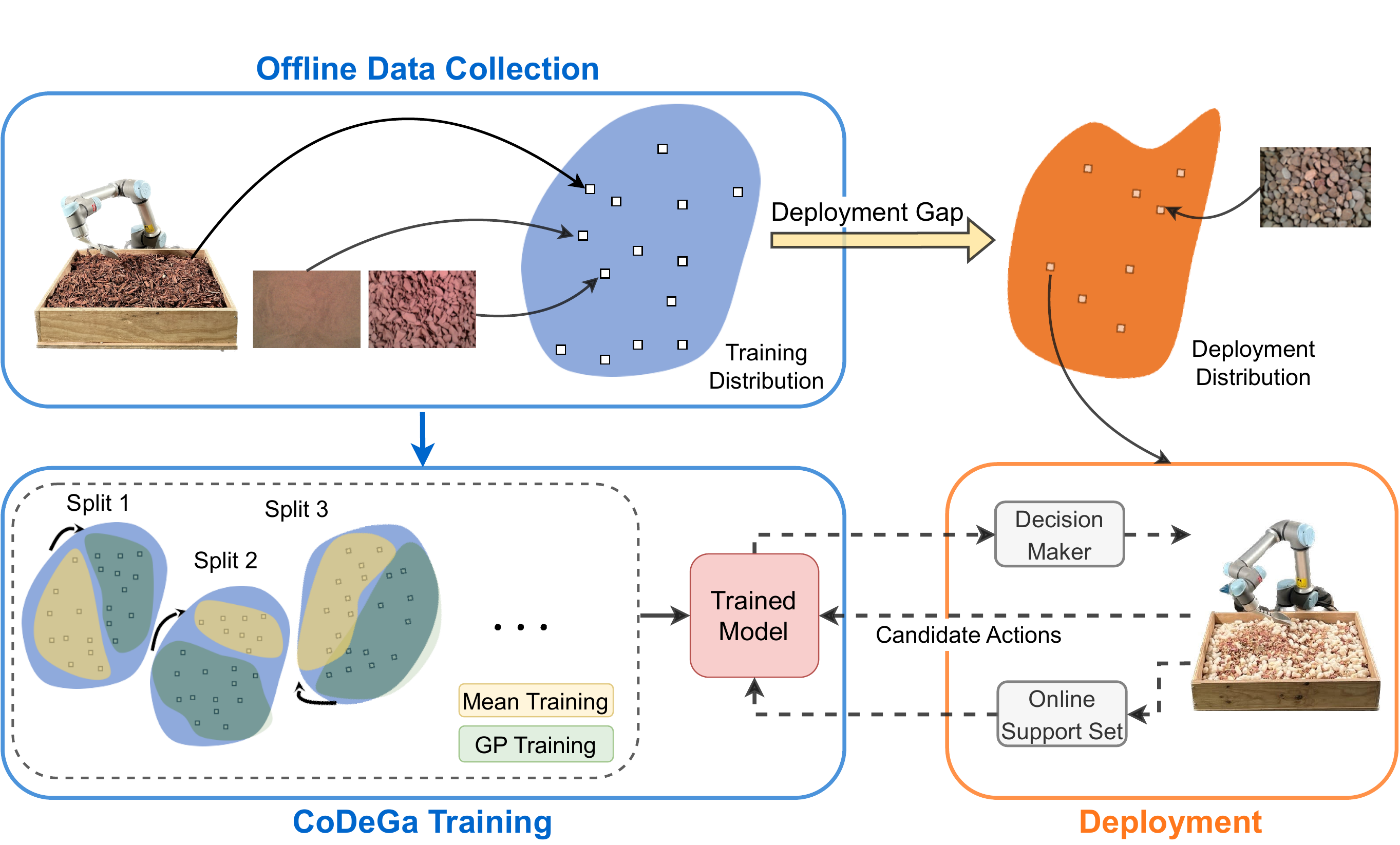}
    \caption{Method overview. Our proposed deep GP model is trained on the offline database with CoDeGa, which repeatedly splits the training set into mean-training and kernel-training and learns kernel parameters to minimize the residuals from the mean models. In deployment, the decision-maker uses the trained model and adapts it to the data acquired online (Support Set).}
    \label{fig:method}
    \vspace{-0.5cm}
\end{figure*}

This paper is inspired by proposed NASA missions to send autonomous landers to Europa and Enceladus to collect and analyze terrain samples to explore whether these bodies exhibit conditions that could support extraterrestrial life~\cite{NASA2017}. However, the composition of the icy regolith is largely unknown and could range from snow-like granules, to flat ice, to complex terrain formations. Although a lander may be tuned to sample well on Earth-bound terrain simulants, it will inevitably face a {\em deployment gap} when operating on an extraterrestrial body. This work proposes a learning-based approach that explicitly addresses such gaps. 

Specifically, we study a simplified but analogous problem of scooping in which the goal is to collect high-volume samples from a target terrain with a limited budget of attempts. Different terrain compositions and shapes require very different scooping strategies, as shown in Fig.~\ref{fig:scooping_strategy}. For instance, while scooping against a slope on sand leads to a larger scooped volume, doing so on a bed of gravel often results in excessive contact forces because the gravel can interlock and jam. Although machine learning models can be trained using terrain simulants to predict scooping outcomes~\cite{Lu2021Excavation}, these models will have worse prediction accuracy when the target terrain is {\em out of distribution} of the training terrains.

As a result, it is necessary for the robot to adapt its strategy in an on-line fashion based on data acquired on the target terrain.  Ideally, the robot should adapt with a small amount of data, i.e., in a few-shot learning setting. {\em Meta-learning}, also known as learning to learn, is a promising approach to few-shot adaptation in which the on-line learning strategy itself is learned using a set of training tasks~\cite{Thrun1998}.

We propose a deep Gaussian process (GP) method that uses meta-learning to learn both the mean and kernel functions. The deep Gaussian process model employs deep mean functions and deep kernels where the input to the GP kernel is transformed by a neural network. It's important to differentiate our definition of the term deep GP from its other usage in the literature where it can also refer to deep belief networks constructed from compositions of GP models ~\cite{damianou2013DeepGP}. Although models similar to our proposed model have been adopted for few-shot adaptation~\cite{Patacchiola2020DKT}, we introduce a new meta-training approach that explicitly trains the kernel to perform well on out-of-distribution tasks. Summarized in Fig.~\ref{fig:method}, our Deep Meta-Learning with Controlled Deployment Gaps (CoDeGa) method repeatedly splits the training set into mean-training and kernel-training and learns kernel parameters to minimize loss over the residuals from the mean models. The key idea is that we explicitly control the splitting process to ensure the mean-training and kernel-training sets have maximal domain gaps, and the residuals seen by the kernels based on these splits are more representative of the residuals seen in an out-of-distribution target task.

During scooping, our model takes an RGB-D image and parameters of a scooping action as input and predicts the mean and variance of the scooped volume.  It is trained by executing 5,100 scoops on a variety of terrains with different compositions and materials. For decision-making, we use a Bayesian optimization framework that chooses an action that maximizes an acquisition function that balances the scoop volume prediction and its uncertainty.  Our experiments evaluate the proposed method on out-of-distribution terrains that have drastically different appearances and/or material properties than the training terrains. Our method allows the robot to achieve high-volume scooping actions in out-of-distribution terrains faster than state-of-the-art deep kernel and conditional neural processes models when used in Bayesian Optimization. Moreover, it significantly outperforms non-adaptive methods such as those proposed in the granular material manipulation literature. Finally, to encourage further research in these domains, we release our dataset of 6,700 executed scoops collected on a diverse set of materials and compositions \footnote{Dataset download link: \url{https://drillaway.github.io/scooping-dataset.html}}

\section{Related Work}
Our work is related to granular material manipulation and is to our knowledge the first approach to integrate vision input and manipulation outcomes to adapt to out-of-distribution terrains. In addition, we also review literature related to the learning techniques underlying our approach, including meta-learning for Gaussian processes, Bayesian optimization, and few-shot learning.

\subsection{Granular Material Manipulation}
Granular materials occur in a variety of real-world robotic applications, including food preparation, construction, and outdoor navigation. Many granular material manipulation tasks have been explored, including scooping~\cite{Schenck2017ScoopingLearning}, excavation~\cite{Dadhich2016ExcavationSurvey}, pushing~\cite{Suh2021Pile}, grasping~\cite{Takahashi2021GraspingFoods}, untangling~\cite{Ray2020Untangling} and locomotion~\cite{Shrivastava2020,Karsai2022}. Most related to our work are scooping and excavation, which are connected but operate on different scales. The task proposed by Schenck et al. focuses on manipulating a granular terrain to a certain shape~\cite{Schenck2017ScoopingLearning} by learning a predictive function of terrain shape change given an action. An optimization-based method is proposed by Yang et al. to generate excavation trajectories to excavate desired volumes of soil based on the intersection volume between the digging bucket swept volume and the terrain~\cite{yang2021optimization}. Dadhich et al. propose to use imitation-learning for rock excavation by wheel loaders, given expert demonstrations of~\cite{Dadhich2016}. All of these past works are developed on a single type of material. In contrast to these methods, our work directly addresses the large deployment gaps that are likely to be found in  extraterrestrial terrain sampling. We explore the use of kernel learning methods trained on offline data combined with Bayesian Optimization to achieve adaptive behavior with little experience of the target terrain.

\subsection{Meta-learning for Gaussian Process and Bayesian Optimization}
When using a Gaussian process (GP) to model an unknown function, knowledge of the distribution from which this function is drawn is required, which is encoded in the GP kernel and other hyperparameters. Usually, this distribution is unknown in practice and all the hyperparameters are estimated with data. However, in the few-shot regime, this estimation can be quite inaccurate with few data points available~\cite{GPML}. One idea to approach this problem is to meta-train the GP on similar tasks~\cite{Bonilla2007,Skolidis2012GPMeta} offline to find good hyperparameters. In addition, to handle high-dimensional inputs, meta-training deep mean functions and/or deep kernels has also been explored~\cite{Fortuin2019, Patacchiola2020DKT,Rothfuss2021PACOH}.

Bayesian Optimization (BO) is a popular approach for sequential optimization in problems like ours that can be modeled as a contextual bandit. Using GP as the Bayesian statistical model for modeling the objective function is common practice in BO. Meta-learning GP in the context of BO has also been explored before, for both GP~\cite{Huang2021BOGPMeta,Wang2018MetaBO} and deep mean and kernels~\cite{Rothfuss2021PACOH}. Closest to our approach is the work that meta-learns deep kernels and means for use in BO~\cite{Rothfuss2021PACOH}. Compared to this work, where there are dozens to hundreds of on-line samples, our work focuses on the few-shot regime. In addition, our work deals with real-world high-dimensional inputs and challenging testing scenarios that are drastically different from training scenarios.

% Using semi-parametric Gaussian process regression is a popular technique for system identification~\cite{Wu2012GP,Camoriano2016GP,Ko2007GP}. 
% Meta-learning acquisition function~\cite{Volpp2020}

\subsection{Few-shot Learning via Meta-learning}
Meta-learning and its application in the few-shot learning scheme have been explored extensively. It involves training models on an offline training dataset consisting of multiple tasks, that can adapt to a novel task using only a few examples. It has been studied in low-dimensional function regression~\citep{Hochreiter2001ICANN,Ha2017ICLR,Ravi2017ICLR}, image classification~\citep{Chen2019ICLR,Snell2017NEURIPS,Vinyals2016NeurIPS,Nichol2018ARXIV}, and reinforcement learning tasks~\citep{Finn2017ICML,Mandi2022ARXIV,Ballou2022ARXIV}. These approaches usually rely on some distance metric in feature space to compare the new examples to the available labeled examples~\citep{Koch2017ICMLW,Santoro2016ARXIV,Vinyals2016NeurIPS,Bartunov2017ICLRW}, perform few-shot estimation of the underlying density of data~\citep{Oord2016NeurIPS,Reed2018ICLR,Bornschein2017NeurIPS,Rezende2016ICML}, or use gradient descent to update a model learned on many related tasks~\citep{Finn2017ICML,Gauch2022ARXIV,Li2017ARXIV,Nichol2018ARXIV}. We apply few-shot learning in the context of decision-making for granular material manipulation. However, these meta-learning methods do not necessarily work well right out of the box for the few-shot adaptive scooping problem, which we will show in the experiments. Hence we propose CoDeGa, a novel few-shot meta-learning method for deep GP that allows the robot to achieve high-quality scooping actions in out-of-distribution terrains faster than state-of-the-art methods.

\section{Problem Formulation}
We formulate the scooping problem as a sequential decision-making task where the robot, in each \textit{episode}, observes the terrain RGB-D image $o \in \mathcal{O}$, uses a \textit{scooping policy} to apply $a \in \mathcal{A}(o)$ where $\mathcal{A}(o)$ is a discrete set of parameterized, observation-dependent scooping motions. The reward $r \in \mathcal{R}$ of a scoop is the scooped volume.
\newline

Presented with a target terrain $T_*$, the robot's goal is to find a scoop whose reward is above a threshold $B$. In planetary missions, for example, $B$ could be the minimal volume of materials needed to perform an analysis. During the $n$-th episode, the robot knows the history of scoops on this terrain $H = \{(o^j,a^j,r^j)\,|\,j=1,\ldots,n-1\}$, which we also refer to as the \textit{on-line support set}. Note that the support set only contains samples of low quality, i.e. below $B$, because otherwise the goal would already have been achieved. 
\newline

The robot has access to an \textit{offline} prior scooping experience, which consists of a set of $M$ terrains $\{T_1, \dots, T_M \}$, and a training dataset $D_i = \{(o^j,a^j,r^j)\,|\,j=1,\ldots,N_i\}$ of past scoops and their rewards for each terrain $i=1,...,M$.

For a terrain, we suppose a \textit{latent variable} $\alpha$ characterizes its composition, material properties, and topography, which are only indirectly observed. Let $\alpha_*$ characterize $T_*$ and $\alpha_i$ characterize $T_i$ for $i=1,\ldots,M$. Moreover, the observation is dependent on the latent variable, and an action's reward $r\equiv r(\alpha,a)$ is also an unknown function of the action and latent variable. Standard supervised learning applied to model $r\approx f(o,a)$ will work well when $\alpha_*$ is within the distribution of training terrains, and $\alpha_*$ is uniquely determined by the observation $o$ or the reward is not strongly related to  unobservable latent characteristics. However, when $T_*$ is out of distribution or the observation $o$ leaves ambiguity about latent aspects of the terrain that affect the reward, the performance of the learned model will degrade. 

As a result, on-line learning from $H$ has the potential to help the robot perform better on $T_*$.  Meta-learning attempts to model the dependence of the reward or optimal policy on $\alpha$, either with explicit representations of $\alpha$ (e.g., conditional neural processes~\cite{garnelo2018CNP}) or implicit ones (e.g., kernel methods~\cite{Patacchiola2020DKT}, which are used here).

\section{Scooping Problem Description}
\begin{figure}[]
\centering
    \includegraphics[width=0.7\linewidth]{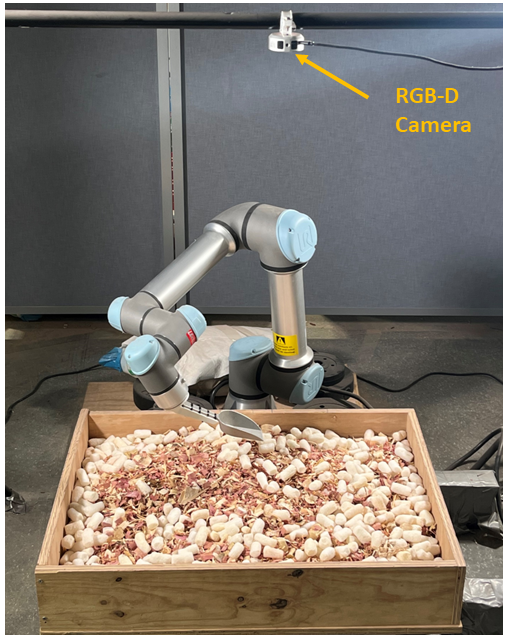}
    \caption{The experimental setup.}
    \label{fig:scooping_setup}
    \vspace{-0.5cm}
\end{figure}

\begin{figure}[]
\centering
    \includegraphics[trim=0cm 2cm 0cm 2cm,width=0.8\linewidth]{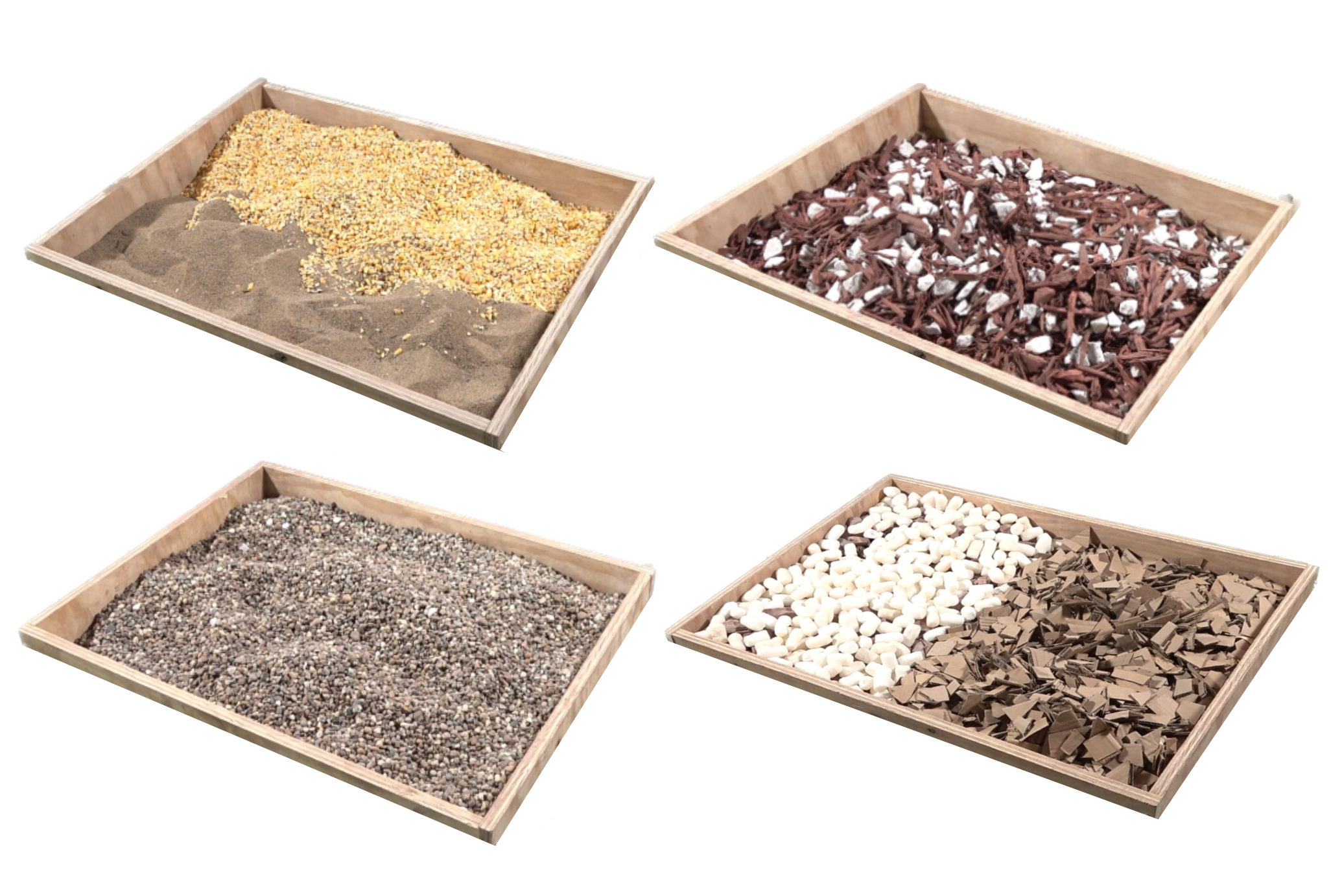}
    \caption{Example terrains illustrating different compositions, materials, and topography. Top left: Partition of Sand and Corn. Top right: Mixture of Mulch and Gravel. Bottom left: Single of Pebbles. Bottom right: Layer of Shredded Cardboard (right), Packing Peanuts (left, top layer), and Slates (left, bottom layer).}
    \label{fig:terrain-examples}
\end{figure}

This section describes the scooping problem in more detail. We use the setup shown in Fig.~\ref{fig:scooping_setup}, which includes a UR5e arm with a scoop mounted on the end-effector, an overhead Intel RealSense L515 RGB-D camera, and a scooping tray that is approximately 0.9\,m x 0.6\,m x 0.2\,m. A terrain is defined as a unique {\em composition} of one or more {\em materials}, where a material is composed of particles with consistent geometry and physical properties. We consider a variety of materials and compositions for the offline database. The materials used in this project are listed in Tab.~\ref{tbl:materials}. The offline database contains materials Sand, Pebbles, Slates, Gravel, Paper Balls, Corn, Shredded Cardboard, and Mulch. The testing terrains also include Rock, Packing Peanuts, Cardboard Sheet and Bedding, which significantly differ from the offline materials in terms of appearance, geometry, density, and surface properties. The terrain compositions used are listed in Tab.~\ref{tbl:compositions}. The offline database contains the Single, Mixture, and Partition compositions, while the testing set also contains the Layers composition. On terrains with the Layers composition, observations do not directly reflect the composition of the terrain, and on-line experience is needed to infer it. All terrains are constructed manually, with varying surface features (e.g. slopes, ridges, etc.) with a maximum elevation of about 0.2\,m and a maximum slope of 30$^{\circ}$. Some terrain examples are demonstrated in Fig.~\ref{fig:terrain-examples}. We also observe that the scooping outcomes show high variance because many terrain properties are not directly observable, such as the arrangement and geometry of the particles beneath the surface.

\begin{table}[h!]
     \begin{center}
     \begin{tabular}{ p{19mm}cp{19mm}c}
     \toprule
      %Name & Appearance & Description & Note \\ 
      %\midrule
   % \cmidrule(r){1-1}\cmidrule(lr){2-2}\cmidrule(l){3-3}
      {\bf Sand}$^\dagger$:
      fine play sand,
      $\ll$ 1\,mm
      & \raisebox{-30 pt}{\includegraphics[width=0.1\textwidth]{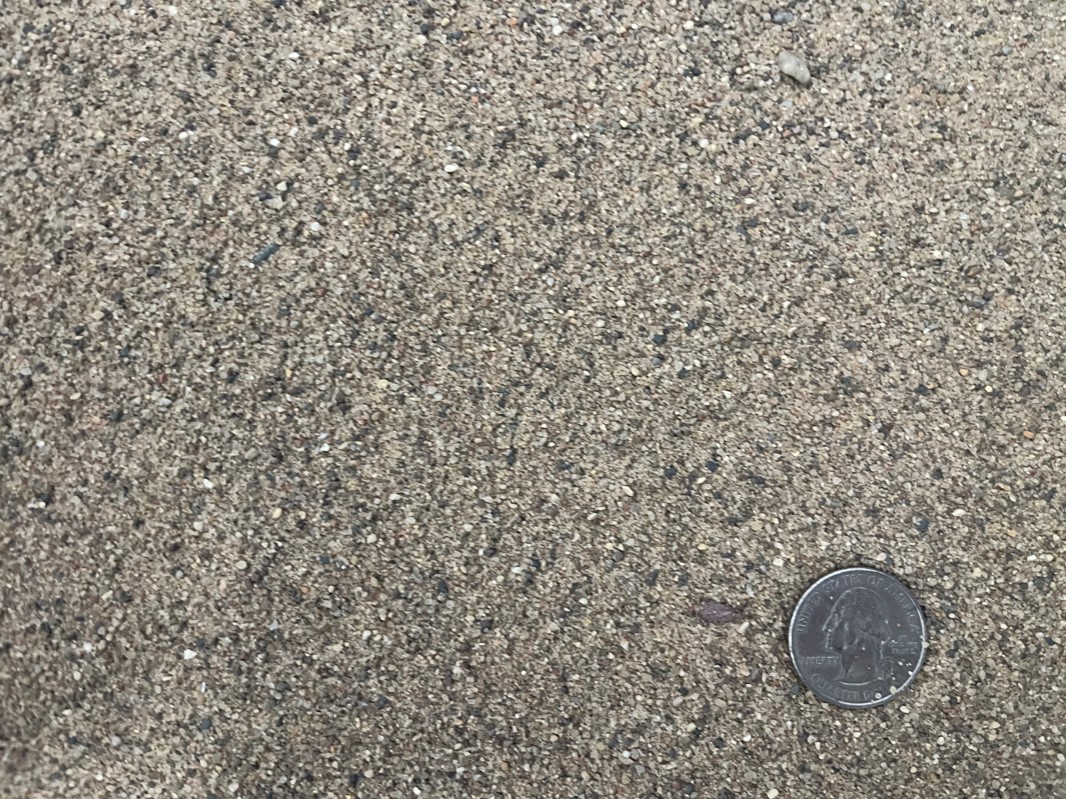}} & 
      {\bf Pebbles}$^\dagger$: rocks, 0.8 -- 1.0\,cm &\raisebox{-30pt}{\includegraphics[width=0.1\textwidth]{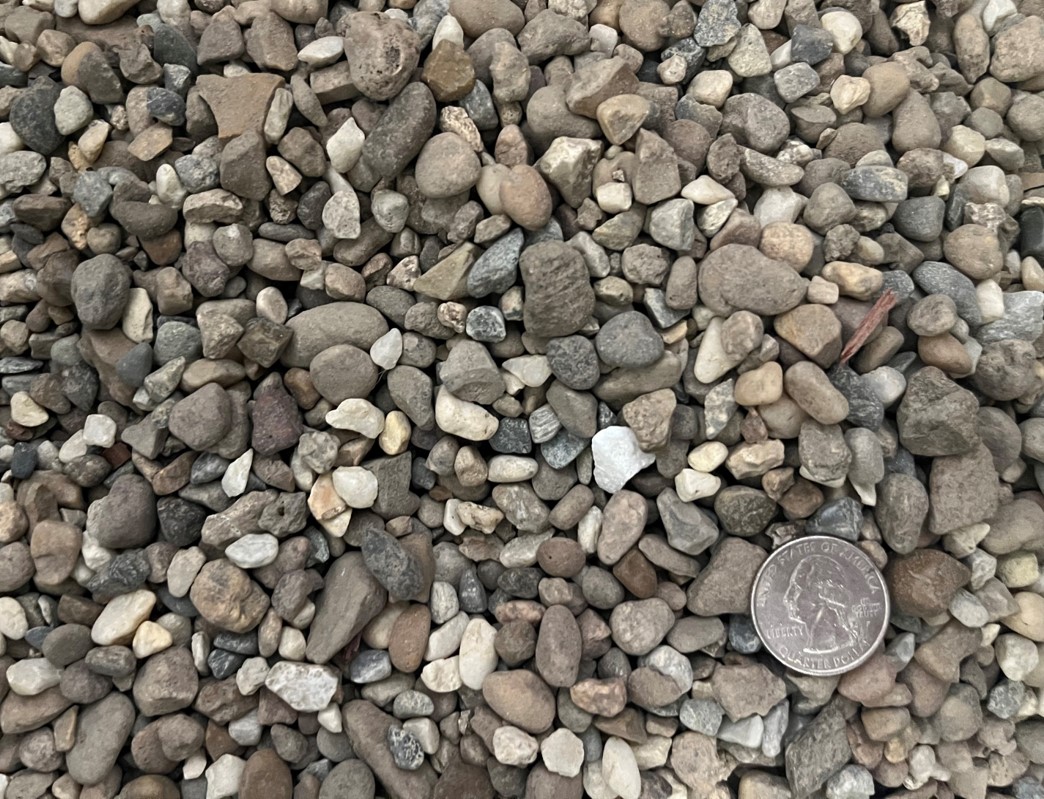}} \vspace{1mm} \\
      %\midrule
      {\bf Slate}$^\dagger$: flat rocks, 2.0--4.0\,cm & \raisebox{-30 pt}{\includegraphics[width=0.1\textwidth]{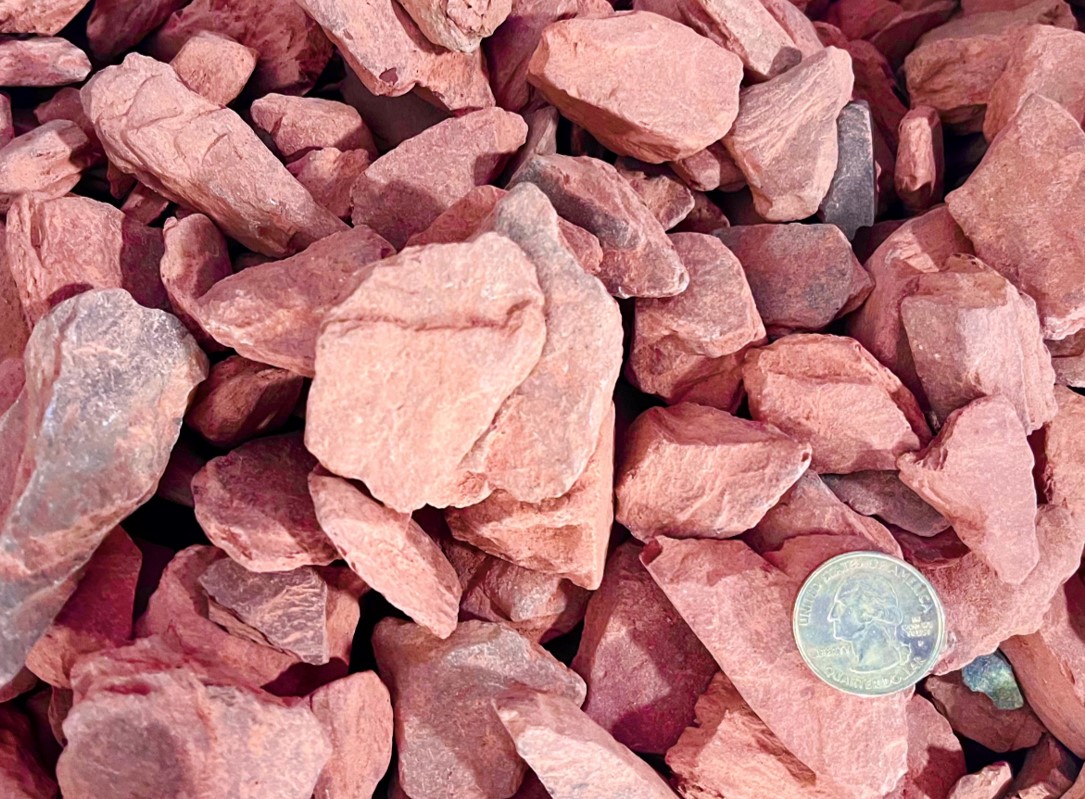}} &
      {\bf Gravel}$^\dagger$: rocks, 1.5--3.0\,cm & \raisebox{-30 pt}{\includegraphics[width=0.1\textwidth]{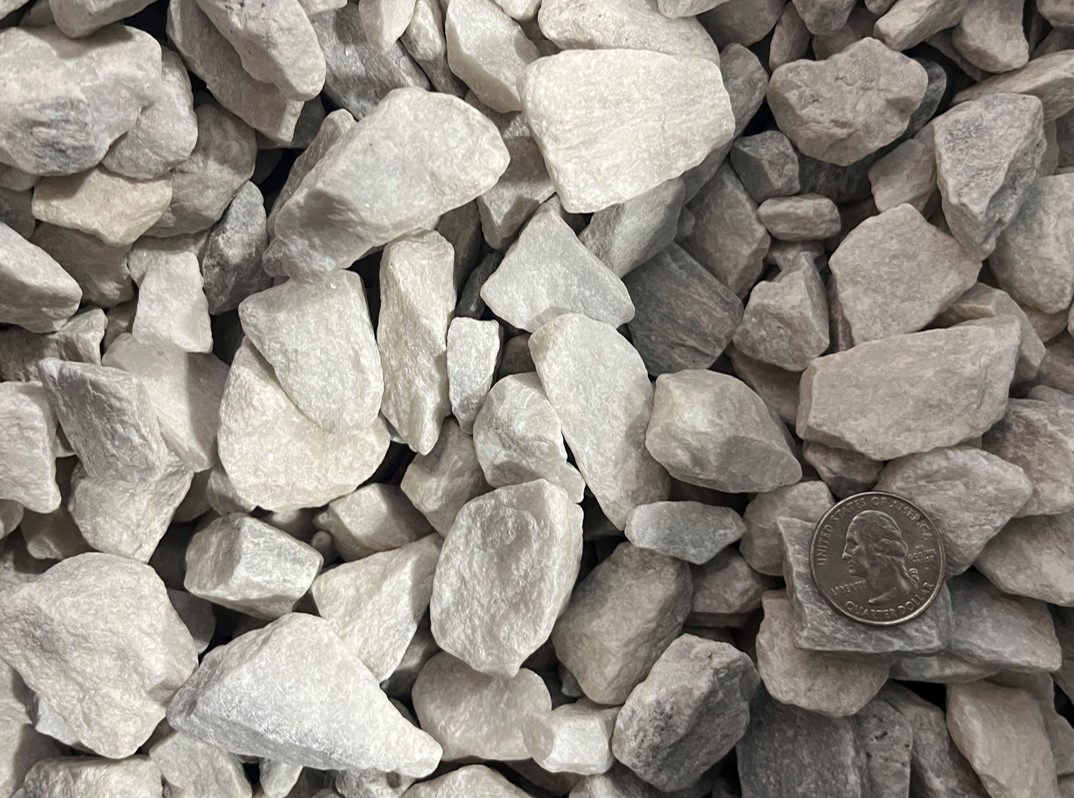}} \vspace{1mm}\\
      %\midrule
      {\bf Paper Balls}$^\dagger$: crumpled paper, 4.0 -- 6.0\,cm & \raisebox{-30 pt}{\includegraphics[width=0.1\textwidth]{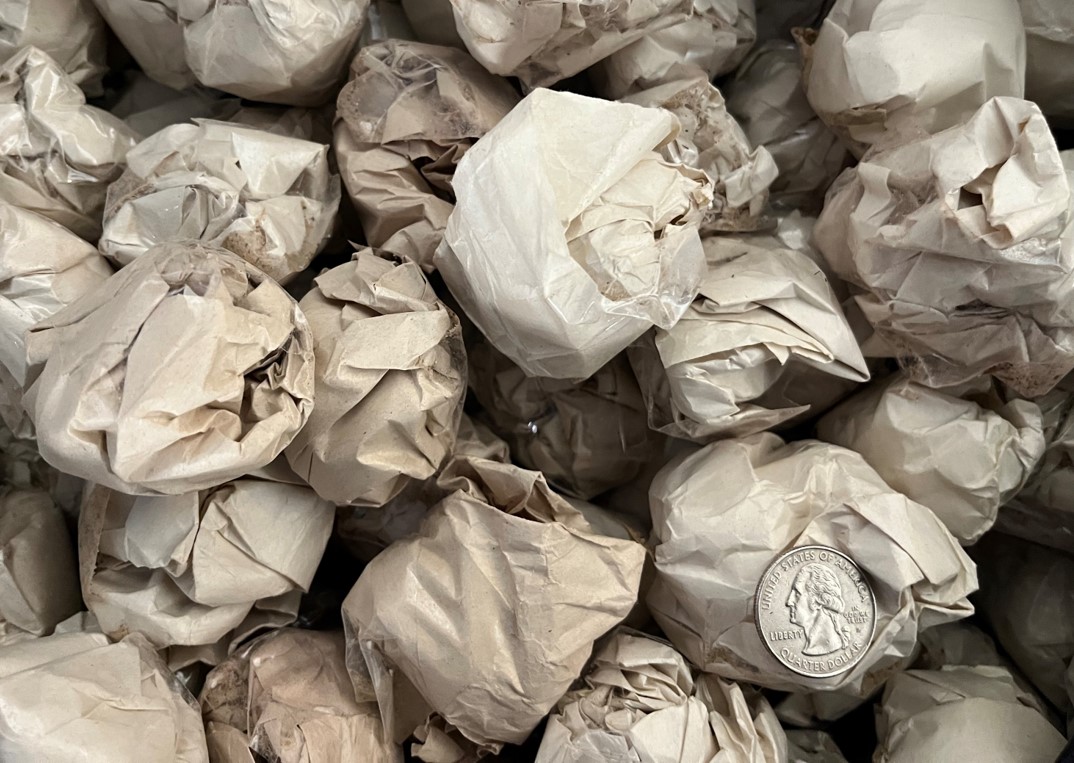}}& 
      {\bf Corn}$^\dagger$: dry corn kernels, 0.3--0.7\,cm & \raisebox{-30 pt}{\includegraphics[width=0.1\textwidth]{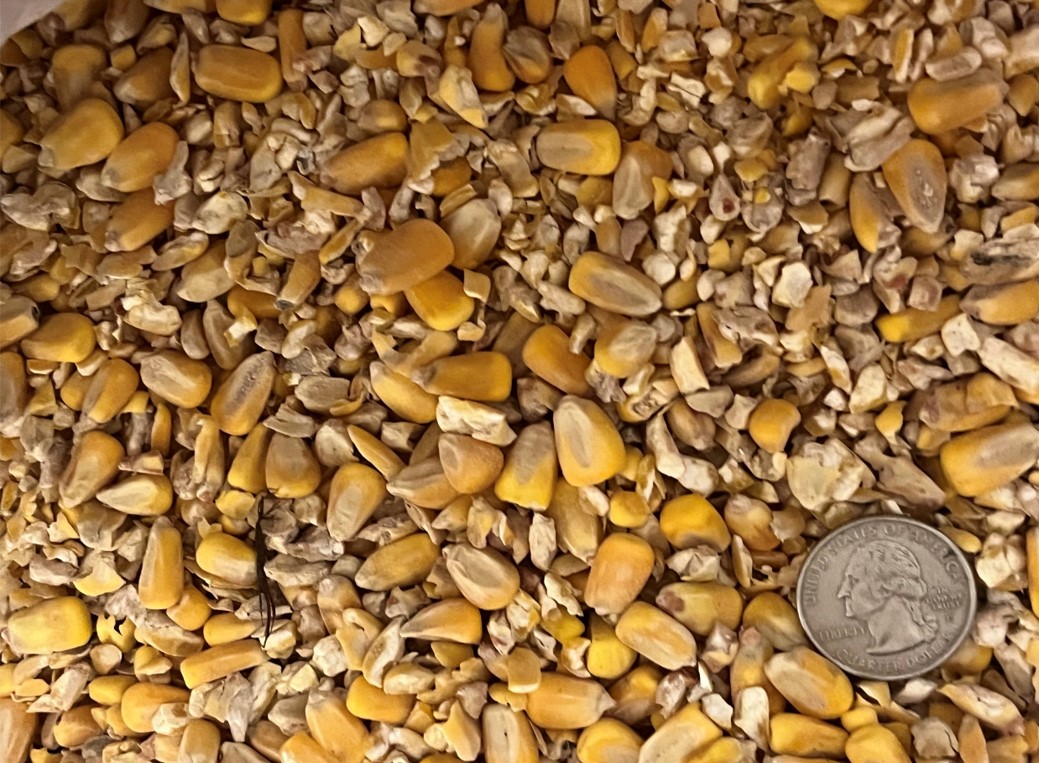}} \vspace{1mm}\\
      %\midrule
      {\bf Shredded Cardboard}$^\dagger$: cardboard, 1.0 -- 8.0\,cm & \raisebox{-30 pt}{\includegraphics[width=0.1\textwidth]{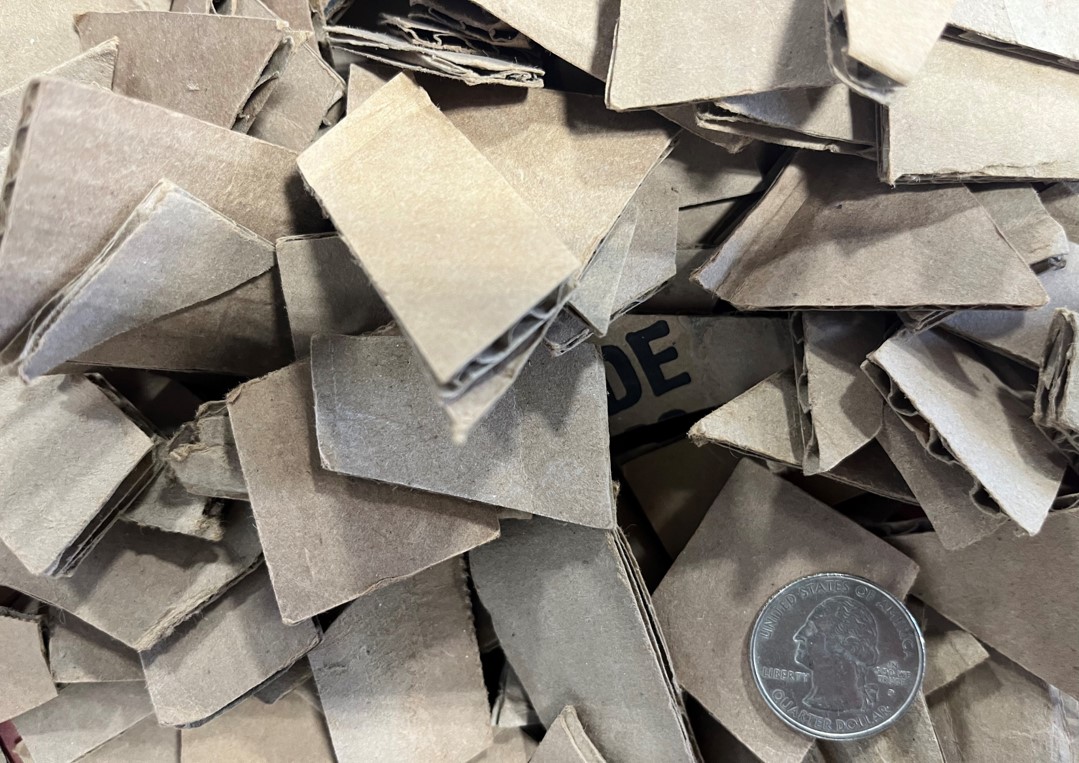}} &
      {\bf Mulch}$^\dagger$: red wood landscape mulch & \raisebox{-30 pt}{\includegraphics[width=0.1\textwidth]{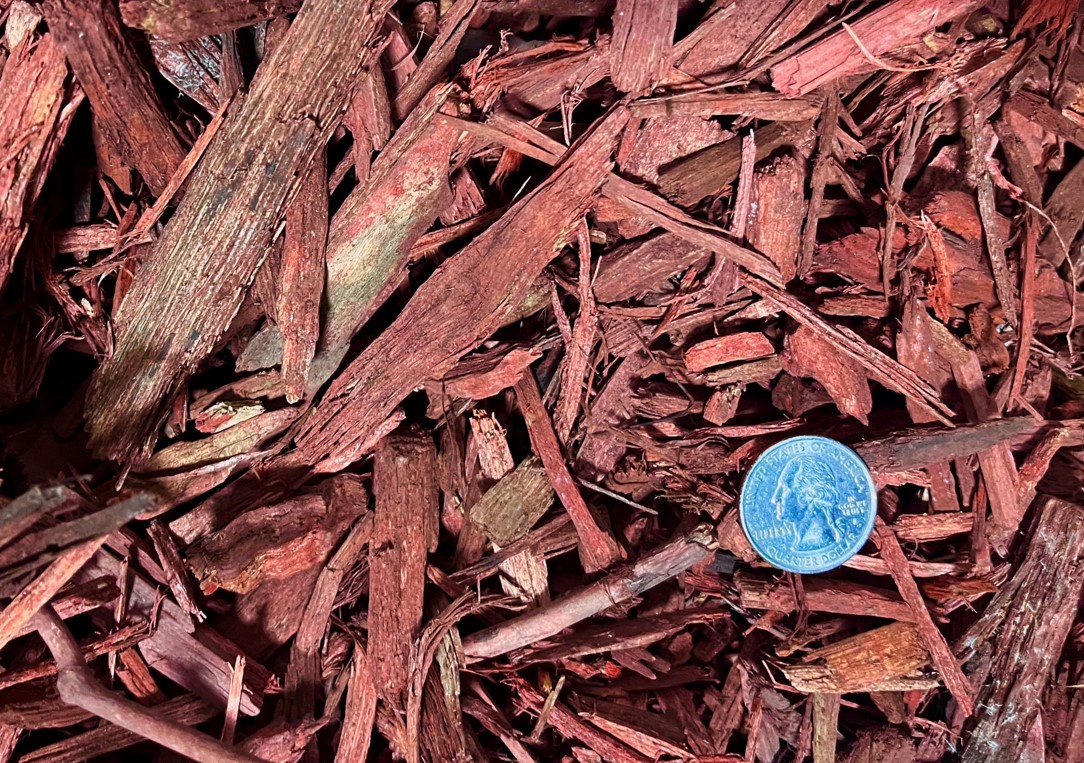}} \vspace{1mm}\\
      %\midrule
      {\bf Rock}: rocks, 5.0 -- 8.0\,cm & \raisebox{-30 pt}{\includegraphics[width=0.1\textwidth]{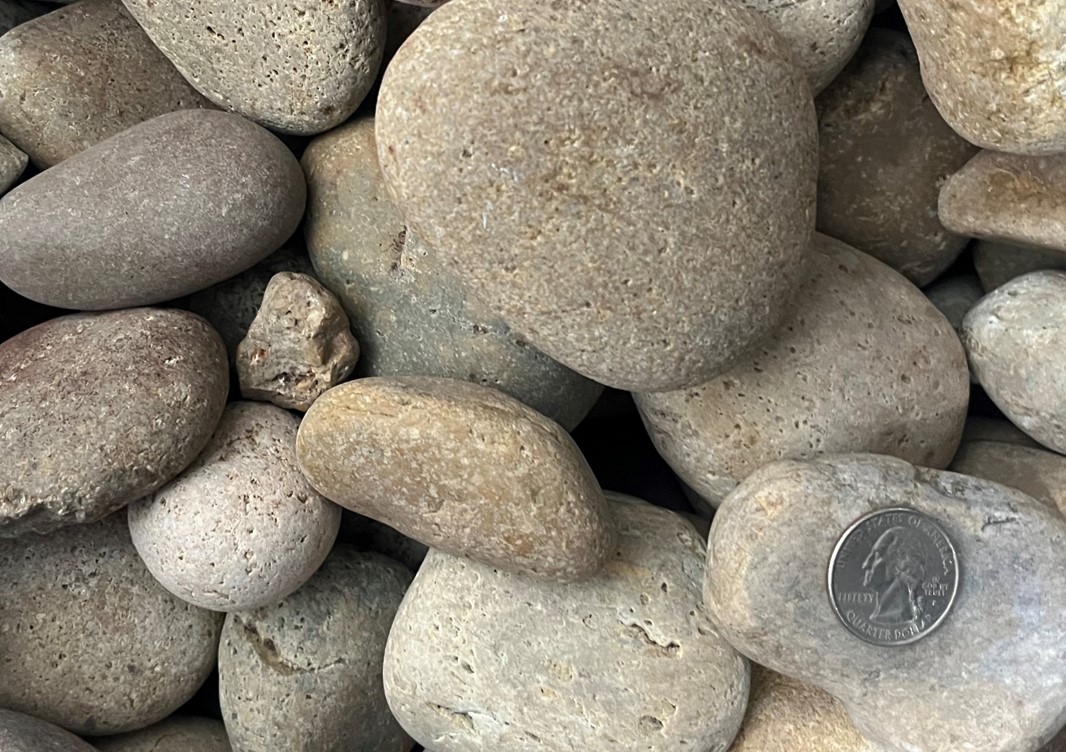}} &
      {\bf Packing Peanuts}: white packing peanuts, 2 x 4\,cm & \raisebox{-30 pt}{\includegraphics[width=0.1\textwidth]{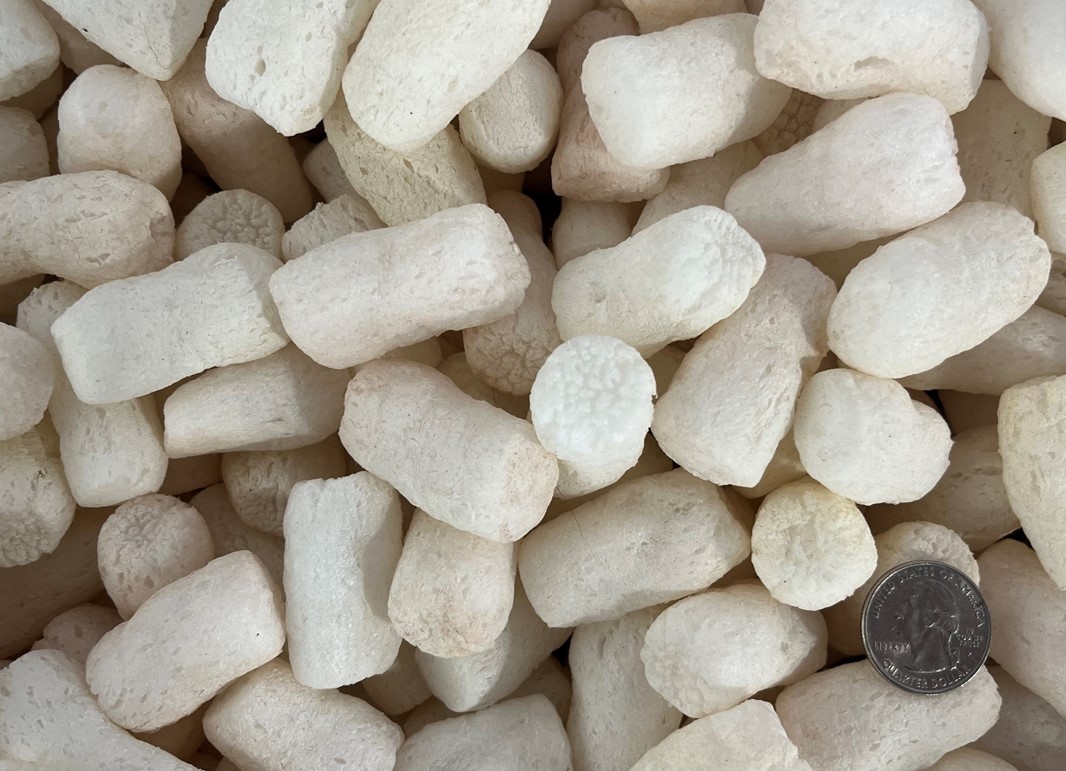}} \vspace{1mm}\\
      %\midrule
      {\bf Cardboard Sheet}: flat cardboard sheet  & \raisebox{-30 pt}{\includegraphics[width=0.1\textwidth]{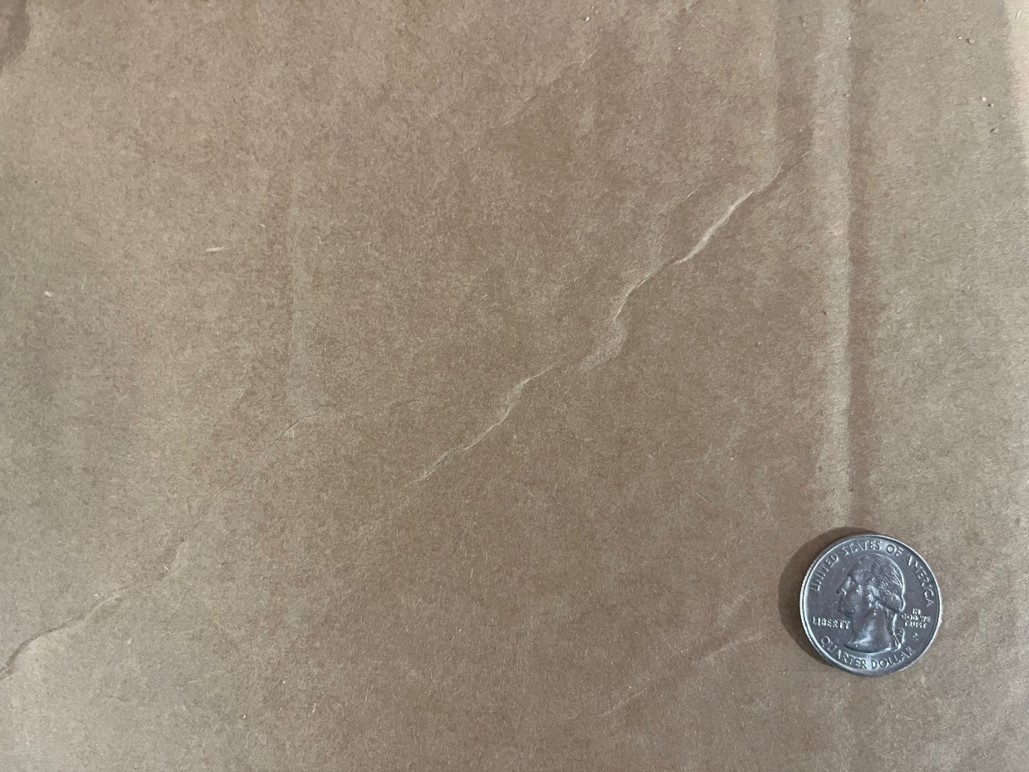}} & 
      {\bf Bedding}: small animal bedding, wood shavings, 0.2 - 3\,cm &\raisebox{-30 pt}{\includegraphics[width=0.1\textwidth]{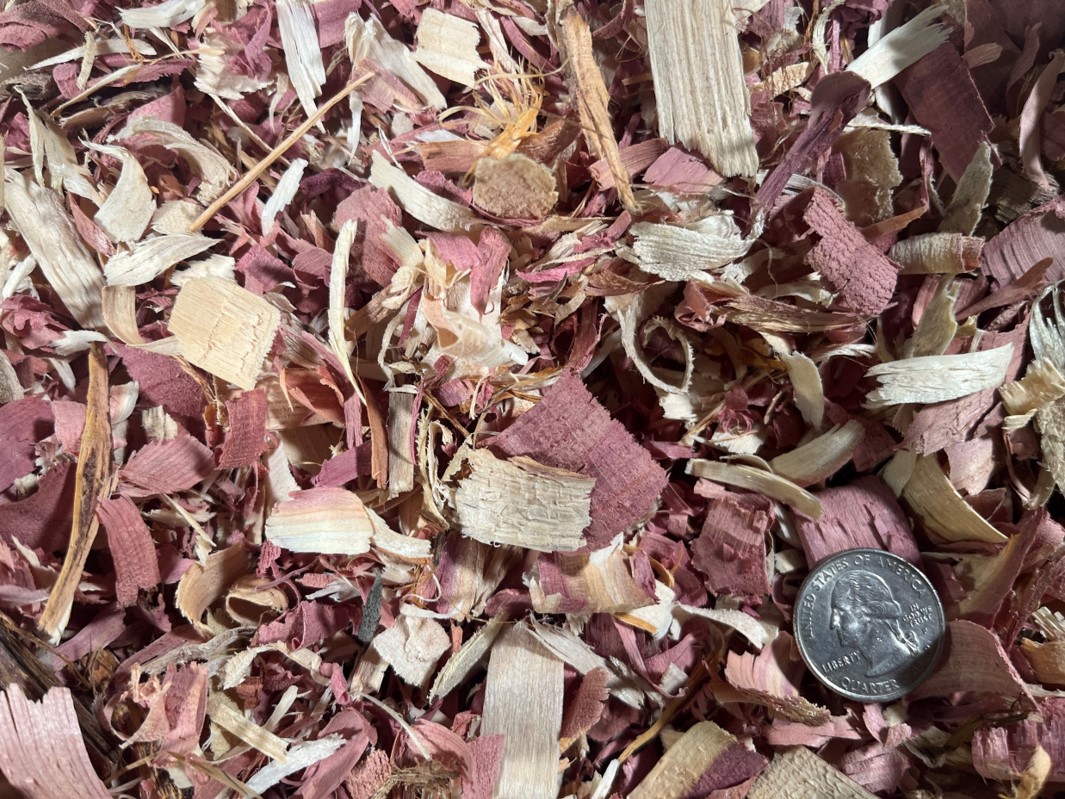}} \\
      \bottomrule
      \end{tabular}
      \caption{Materials used in each terrain, with approximate grain sizes where applicable. Images show U.S. quarter coin for scale. Training set denoted with $^\dagger$; testing set consists of all materials and novel material combinations.}
      \label{tbl:materials}
      \end{center}
  \end{table}

\begin{table}[h!]
     \begin{center}
     \begin{tabular}{ p{21mm}cp{21mm}c}
     \toprule
      %Name & Appearance & Description & Note \\ 
      %\midrule
   % \cmidrule(r){1-1}\cmidrule(lr){2-2}\cmidrule(l){3-3}
      {\bf Single}$^\dagger$:
      A single material
      & \raisebox{-20 pt}{\includegraphics[width=0.08\textwidth]{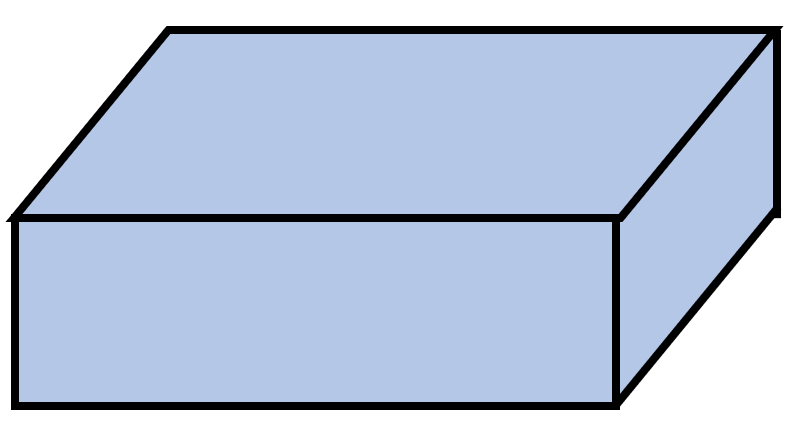}} & 
      {\bf Mixture}$^\dagger$: Uniform mixture of two materials & \raisebox{-20 pt}{\includegraphics[width=0.08\textwidth]{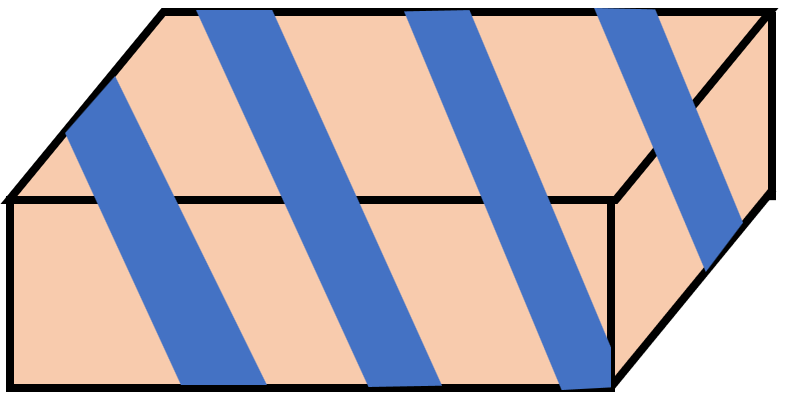}} \vspace{1mm}
       \\
      %\midrule
      {\bf Partition}$^\dagger$: two materials that each occupy a partition 
      &\raisebox{-20pt}{\includegraphics[width=0.08\textwidth]{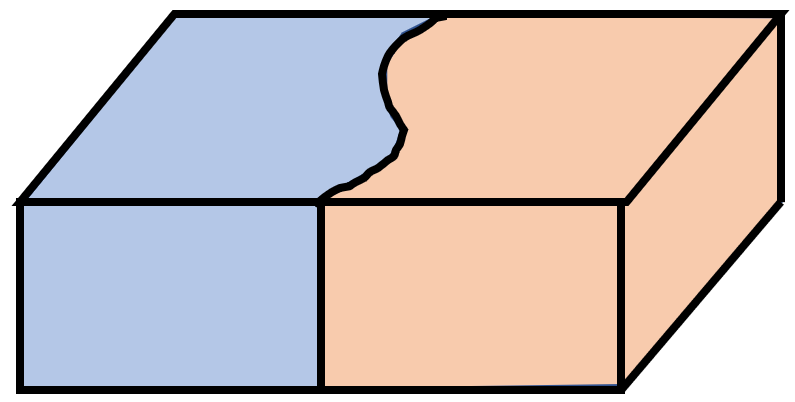}} 
       &
      {\bf Layers}: Two partitions but with two layers of different materials in one partition & \raisebox{-20 pt}{\includegraphics[width=0.081\textwidth]{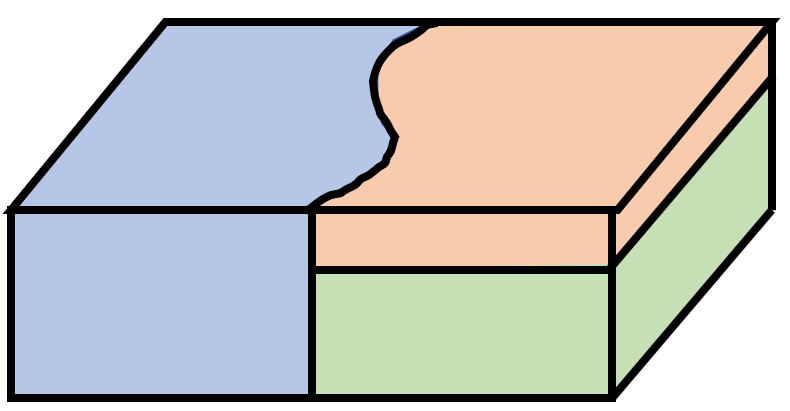}} \vspace{1mm}\\

      \bottomrule
      \end{tabular}
      \caption{Terrain compositions. Training set denoted with $^\dagger$; testing set consists of all compositions.}
      \label{tbl:compositions}
      \end{center}
      \vspace{-0.5cm}
  \end{table}

\begin{figure}[]
    \includegraphics[trim=0.5cm 6cm 10.5cm 2.1cm,clip,width=\linewidth]{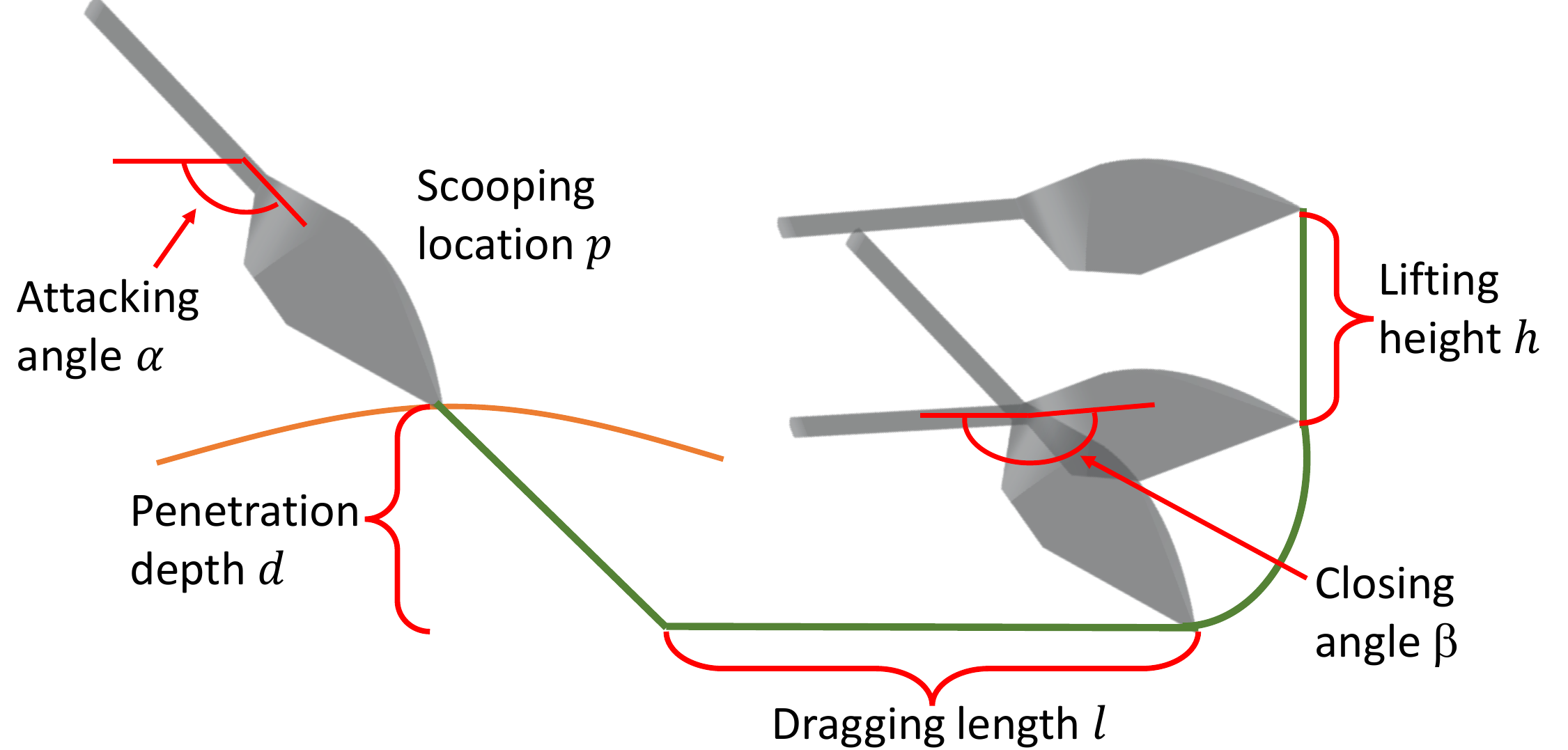}
    \caption{Scooping trajectory.}
    \label{fig:trajectory}
\end{figure}

A \textit{scoop action} is a parameterized trajectory for a scoop end effector that is tracked by an impedance controller. We follow the common practice in the excavation literature~\cite{Lu2021Excavation,sing1995synthesis} to define a scooping trajectory, shown in Fig.~\ref{fig:trajectory}, where the scoop has a roll angle of 0 and stays in a plane throughout the trajectory. The scoop starts the trajectory at a location $p$, penetrates the substrate at the attack angle $\alpha$ to a penetration depth of $d$, drags the scoop in a straight line for length $l$ to collect material, closes the scoop to an angle $\beta$, and lifts the scoop with a lifting height $h$. We assume that the scoop always starts scooping at the terrain surface, which can be calculated from the depth image. The impedance controller of the end-effector is configured with stiffness parameters $b$.

To reduce the action space, we manually tuned the parameters that have smaller effects on the scooping outcome, fixing the attacking angle $\alpha$ at 135$^{\circ}$, the dragging length $l$ at 0.06\,m, the closing angle $\beta$ at 190$^{\circ}$, and the lifting height $h$ at 0.02\,m. In addition, we set two options for the impedance controller stiffness $b$, corresponding to soft and hard stiffness, where the linear spring constants are 250\,N/m and 750\,N/m and the torsion spring constants are 6\,Nm/rad and 20\,Nm/rad, respectively. Therefore, the action is specified by the starting $x$, $y$ position and yaw angle of the scoop, the scooping depth $d$, and stiffness $b$.

To measure the scooped volume, the scoop is moved to a fixed known pose, after which a height map within the perimeter of the scoop is obtained from the depth image. The volume is then calculated by integrating the difference between this height map and the height map of an empty scoop at the same pose collected beforehand.

\begin{figure}[]
\centering
    \includegraphics[width=0.8\linewidth]{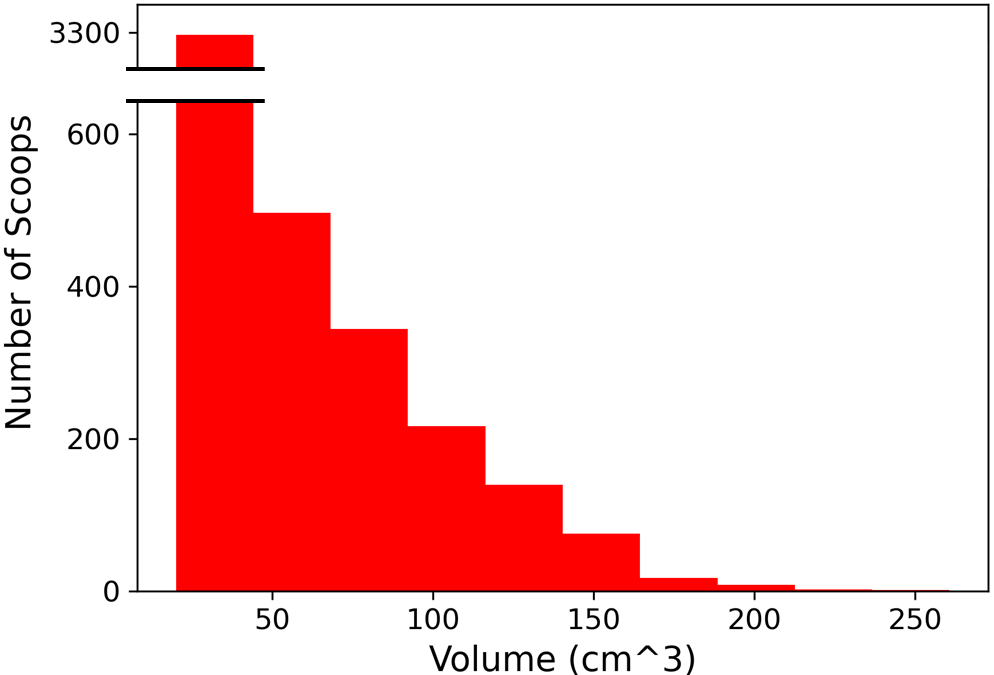}
    \caption{Collected data distribution. The maximum scoop volume is 260.8\,cm$^3$, while the average is 31.3\,cm$^3$. }
    \label{fig:volumes}
    \vspace{-0.5cm}
\end{figure}

The offline database contains data on 51 terrains, all with unique combinations of materials and compositions. Out of these terrains, 8 are Single, 25 are Partition, and 18 are Mixture. The materials used are randomly selected from the training materials. 100 random scoops are collected on each terrain, sampled uniformly with random $x$, $y$ positions in the terrain tray, random yaw angle from a set of 8 discretized yaw angles, 45$^\circ$ apart, random depth in the range of 0.03\,m to 0.08\,m, and random stiffness (either ``hard'' or ``soft''). Sometimes trajectory planning of the robot manipulator for a sampled scoop can fail due to kinematic constraints. If so, the scoop action is discarded and sampling continues until planning is successful. The average scooped volume across the offline database is 31.3\,cm$^3$, and the maximum volume is 260.8\,cm$^3$. The distribution of scooped volumes in the database is shown in Fig.~\ref{fig:volumes}.

\section{Proposed Method}

Our approach models the reward's dependence on the observation $o$, action $a$, and history $H$ as a deep Gaussian process (GP) model.  Our Deep Meta-Learning with Controlled Deployment Gaps (CoDeGa) method uses a deep mean and kernel for the GP which are meta-trained  to perform well under simulated deployment gaps extracted from the training set.  With such a model, the predicted reward and its variance are used at each step to optimize the chosen action using Bayesian optimization. We will first describe our proposed deep GP model, and then training the model with CoDeGa.

For convenience of notation, we let $x=(o,a)$ denote an observation-action pair, and $y=r$ denote a reward. Let us also separate the training datasets into sequences of dependent variables $D_i^y = \{(r_i^j)\,|\,j=1,\ldots,N_i\}$ and independent variables $D^x_i = \{(o_i^j,a_i^j)\,|\,j=1,\ldots,N_i\}$. 

% Similarly, let us separate the on-line history $H$ into $D_*^y = \{(r^j)\,|\,j=1,\ldots,n-1\}$ and $D_*^x = \{(o^j,a^j)\,|\,j=1,\ldots,n-1\}$.

\subsection{Deep Gaussian Process Model}
A GP models function $f$ as a collection of random variables $f(x)$ which are jointly Gaussian when evaluated at locations $x$~\cite{GPML}. A GP is fully specified by its mean function $m(\cdot)$ and kernel $k(\cdot,\cdot)$, which is the covariance function:
\begin{equation}
    f(x) \sim \mathcal{GP}(m(x),k(x,x')).
\end{equation}

Given $n$ existing observed function values $\mathbf{y} = [y_1,\dots,y_n]^T$ at $\mathbf{x} = [x_1,\dots,x_n]^T$, GP regression predicts the function values at new point $x^*$ as a Gaussian distribution:
\begin{equation}
\begin{split}
    P(y^*|&\mathbf{x}, \mathbf{y}, x^*) \sim \\
    &\mathcal{N}(m(x^*) + \mathbf{k}\mathbf{K}^{-1}\bar{\mathbf{y}},k(x^*,x^*) - \mathbf{k}\mathbf{K}^{-1}\mathbf{k}^T ).
\end{split}
\end{equation}
Here,
\begin{equation*}
\begin{split}
        &\mathbf{K} = \begin{bmatrix}
                k(x_1,x_1) & \cdots & k(x_1,x_n)\\
                \vdots & \ddots & \vdots\\
                k(x_n,x_1) & \cdots & k(x_n,x_n)\\
                \end{bmatrix} + \sigma_n^2\mathbf{I},  \\
        &\mathbf{k} = \begin{bmatrix}
            k(x^*,x_1), & \cdots, & k(x^*,x_n)\\
            \end{bmatrix},  \\
        &\bar{\mathbf{y}} = [y_1 - m(x^*),\dots,y_n - m(x^*)]^T,
\end{split}
\end{equation*}
where $\sigma_n$ is the standard deviation of noise at an observation and $\bar{\mathbf{y}}$ is the residual. A typical choice for the mean function is a constant mean and the radial basis function kernel (RBF) is a popular kernel of choice~\cite{GPML}. The mean constant, kernel function parameters, and $\sigma_n$ can be hand-picked if there is prior knowledge of $f$. In practice, doing so is usually not possible and they are estimated from data with type-II maximum likelihood by minimizing the negative log marginal likelihood (NLML):
\begin{equation}
\begin{split}
    -\log P(&
    \mathbf{y}|\mathbf{x}, \theta) = \frac{1}{2} \log |\mathbf{K} + \sigma_n^2 \mathbf{I}| \\
    &+ \frac{1}{2}(\mathbf{y} - m(\mathbf{x}))^T(\mathbf{K} + \sigma_n^2 \mathbf{I})(\mathbf{y} - m(\mathbf{x})) \\
    &+ c,
    \end{split}
\label{eq:LML}
\end{equation}
where $\theta$ denotes all the parameters to be determined and $c$ is a constant.

\begin{figure}[]
\centering
    \includegraphics[trim=4cm 7cm 4cm 8.5cm,clip,width=\linewidth]{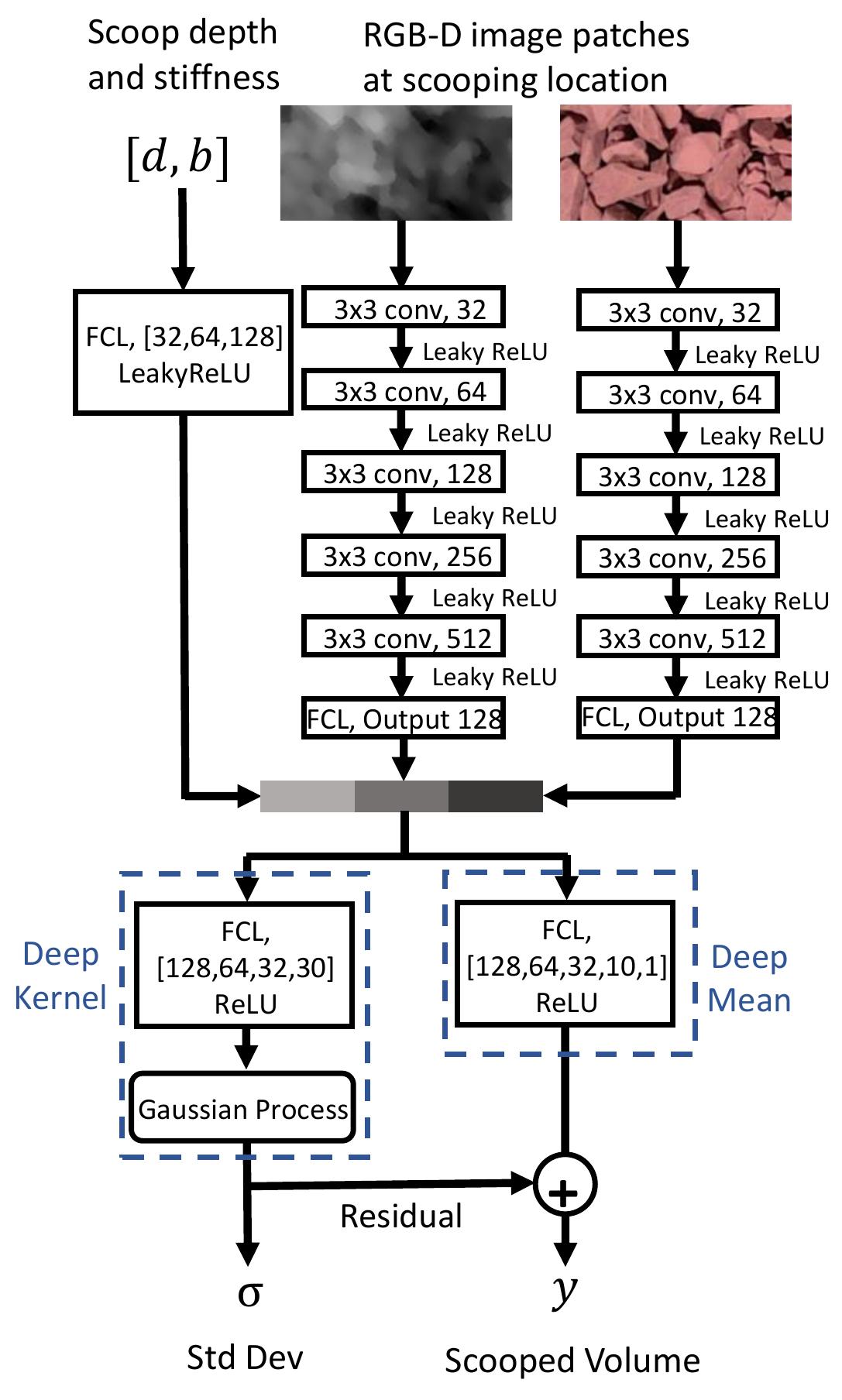}
    \caption{Model architecture. The deep kernel and the deep mean share a common feature extractor.``FCL" denotes fully-connected layers, with the output layer sizes and the activation function specified. ``conv" denotes convolution layers with filter size and number specified. A stride of 2 is used for all convolutional filters. The radial basis function kernel is used for GP.}
    \label{fig:model}
    \vspace{-0.5cm}
\end{figure}

Deep kernels enhance kernels with neural networks to be more scalable and expressive~\cite{wilson2016DKL}. For deep kernels, an input vector is mapped to a latent vector using a neural network before going into the kernel function $k(g_\theta(\cdot), g_\theta(\cdot))$, where $g_\theta(\cdot)$ is a neural network with weights $\theta$.  Deep kernels for the few-shot adaptation setting have been proposed~\cite{Patacchiola2020DKT} previously. Deep kernels were also extended to use deep mean functions $m_\theta(\cdot)$~\cite{Fortuin2019}. Our proposed deep GP contains both a deep kernel and a deep mean. The model takes in a local patch of the RGB-D image starting at the sampling location and aligned with the yaw angle and the action parameters to predict the scooped volume. Since the information on the scooping starting location and yaw angle is already contained in the image patch, the action parameters include only the scooping depth and the binary stiffness variable. Note that we do not use the entire image because an image patch at the scoop location contains most of the information needed to evaluate a scoop, and is much more computationally efficient. The model architecture is shown Fig.~\ref{fig:model}. The kernel and mean function share the same feature extractor, which is a convolutional neural network, and have separate fully connected layers. 

\subsection{Meta-learning with Controlled Deployment Gaps}
The neural network parameters and the kernel parameters of a deep GP can be jointly trained over the entire training set with the same NLML loss as Eqn.~\ref{eq:LML}, where $\theta$ contains the neural network parameters.

However, this approach does not typically train kernels that are well-tuned to individual tasks because it aggregates the data from all tasks together. Instead, meta-training may be realized with stochastic gradient descent with each batch containing the data for a single task, i.e. minimizing the aggregate loss $\sum_i -\log P(D_i^y|D_i^x, \theta)$, where $D_i^y$ and $D_i^x$ are the target variables and input variables of a single task.  We will refer to this approach as Deep Kernel and Mean Transfer (DKMT) which has been proposed in the meta-learning literature~\cite{Fortuin2019, Rothfuss2021PACOH}.

DKMT has potential problems with out-of-distribution tasks. During training, the residuals seen by the kernels are residuals of the deep mean on the training tasks, which could be very different compared to the residuals on tasks out of distribution of the training terrains. As our experiments will show, this feature leads to the kernels being poorly calibrated in the worse case. Another potential issue is the over-fitting of the deep mean function. The first two terms of NLML in Eqn.~\ref{eq:LML} are often referred to as the \textit{complexity penalty} and \textit{data fit} terms, where the \textit{complexity penalty} regularizes the deep kernels~\cite{GPML}. However, there is no regularization of the deep mean function. As a result, the deep mean can potentially overfit on all the training data, so the residuals will be close to zero.  

% Even when the deep mean is regularized or its capacity is limited, the residuals tend toward noise which makes the kernel less likely to generalize.

CoDeGa addresses these issues by encouraging the residuals seen in kernel training to be representative of the residuals seen in out-of-distribution tasks with novel materials. The idea is to separate the training terrains into a mean training set and a kernel training set, where each set contains different materials from one another. Then, the mean is trained on the mean training set to minimize error and the GP is trained on the residuals of the mean model on the kernel training set. These residuals are representative of those during deployment because they are characterized by a deployment gap of novel materials. However, this limits the amount of training data available for the mean and kernel. Therefore, we repeat this process similarly to $k$-fold cross-validation, in which each fold has a separate mean model trained on the mean split for that fold, and then the residuals for that model on the kernel split are used to define the kernel loss for that fold. A common kernel is trained using losses aggregated across folds. Finally, the mean model is trained again on all data.

The overall CoDeGa training procedure is as follows. Denote the set of all training materials as $A$ and the set of training datasets as $D=\{D_1,...,D_M\}$.  
\begin{enumerate}[leftmargin=*]
    \item Split $A$ into $K$ folds $A_1,...,A_K$.
    \item Initialize an empty residual training dataset $E$.
    \item For each fold $A_k$:
    \begin{enumerate}
        \item Split $D$ into a kernel training set $S^k_{kernel}$ and a mean training set $S^k_{mean}$. $S^k_{kernel}$ corresponds to terrains that contain any material in $A_k$, regardless of the composition: this means that $S^k_{kernel}$ can contain terrains with materials out of $A_k$ for multi-material compositions. $S^k_{mean}$ consists of $D\setminus S^k_{kernel}$.
        \item Train feature extractor weights $\theta_f^k$ and deep mean weights $\theta_m^k$ using a mean squared error (MSE) loss on all the data in $S^k_{mean}$. Let these weights be denoted $\theta^k$.
        \item For each dataset $D_i \in S^k_{kernel}$, collect the residuals of the mean model $\tilde{r}_i^j = r_i^j - m_{\theta^k}(x_i^j)$, $j=1,...,N_i$. Construct the inputs $\hat{D}_i^x = D_i^x$ and the outputs $\hat{D}_i^y = \{\tilde{r}_i^j\,|\,j=1,...,N_i\}$ to predict the residuals.  $E \leftarrow E \cup \hat{D}_i$.
    \end{enumerate} 
    \item Train deep kernel parameters $\theta_k$ with database $E$, i.e. minimizing the aggregate NLL loss $- \sum_i \log P(\hat{D}_i^y|\hat{D}_i^x, \theta)$. We use stochastic gradient descent where each batch contains all samples in one task. (Note that the associated features extractor weights for each task need to be loaded and fixed at the start of each batch training. )
    \item Train $\theta_f$ and $\theta_m$ from scratch using standard supervised learning on all data in the offline database.
    
\end{enumerate}

The CoDeGa approach could be generalized outside of the scooping domain by noting that material-based splitting intends to maximize the gap between the tasks in the mean / kernel split. If there are known features of tasks that are correlated with the latent task variables $\alpha$, one approach would be to perform clustering in the feature space to identify large splits. We hope to explore these avenues in future work.

\subsection{Bayesian optimization decision-maker}

To use a reward model in the scooping sequential decision-making problem, the decision-maker maximizes a score $s(o,a)$ over the action $a$: $\pi(o) = \arg \max_{\mathcal{A}(o)} s(o,a)$.  A greedy optimizer would use the mean as the score, $s(o,a) = m(o,a,H)$ where $m(o,a,H)=E[r | r \sim p(R\,|\,o,a,H)]$, but this does not adequately explore actions for which the prediction is uncertain. Instead, a Bayesian optimizer uses an acquisition function that also takes uncertainty into account. For example, the upper confidence bound (UCB) method defines the acquisition function $s_{UCB}(o,a)=m(o,a,H) + \gamma \cdot \sigma(o,a,H)$, where $\sigma(o,a,H)=Var[r | r \sim p(R\,|\,o,a,H)]^{1/2}$ is the standard deviation of the prediction and $\gamma > 0$ is a parameter that encourages the agent to explore actions whose outcomes are more uncertain.

% \begin{algorithm}[h]
% \SetAlgoLined
% \textbf{Input:} All terrain data $\{D_i\}$, with their associated terrain materials; set of all terrain materials $A$; number of folds $K$, deep kernel with deep mean model $M$ ($\theta_f$, $\theta_m$, and $\theta_k$ represent the feature extractor, deep mean, and deep kernel parameters)\;
% Train $\theta_f$ and $\theta_m$ on $\{D_i\}$ with standard supervised learning and save to disk\;
% Split $A$ into $K$ folds\; 
% \tcp{Collect mean residuals on simulated deployment gaps}
% Initialize empty residual data $E \leftarrow \emptyset$ \;
% \For{ fold $k_j$ in the $K$ folds}{
% Split the terrains into ones that contain materials in $k_j$ and those that do not: $S_1$ and $S_2$ \;
% Retrain $\theta_f$ and $\theta_m$ from scratch on $S_2$ with standard supervised learning\;
% Save $\theta^j_f$ to disk\;
% \For{$D_i$ in $S_1$}
% {Calculate mean function residuals and build dataset $\hat{D}_i$ with residuals instead of volume\;
% $E \leftarrow E \cup \hat{D}_i$\;}}
% \tcp{Meta-training deep kernels}
% Train $\theta_k$ with the NLML loss on each task, with $\theta^j_f$ associated with $\hat{D}_i$ from disk\;
% Reload $\theta_f$ and $\theta_m$ from disk\;
%  \Return  $M$\;
%  \caption{Meta-training with Controlled Deployment Gaps}
%  \label{alg:training}
% \end{algorithm}

\section{Experiments and Results}
\subsection{Testing Tasks}
We evaluate our method on 16 test terrains that contain out-of-distribution materials and compositions. We introduce 4 new materials, which are Rock, Packing Peanuts, Cardboard Sheet, and Bedding, described in Tab.~\ref{tbl:materials}. In addition to the compositions during training, we also consider the Layers composition, described in Tab.~\ref{tbl:compositions}. Note that the Cardboard Sheet material is not scoopable. For each of the Single, Partition, Mixture, and Layers compositions, we consider 4 terrains. The 4 Single terrains are created with each of the 4 new testing materials. Material combinations on terrains with the Mixture, Partition, and Layers compositions are randomly generated but with the constraints that 1) each of the new  materials is selected at least once; 2) each terrain contains at least 1 new material. We exclude Cardboard Sheet from Mixture since it is physically impossible to create. 

\subsection{Model Training}
PyTorch~\cite{PyTorch} and GPyTorch~\cite{gardner2018gpytorch} are used to implement the neural networks and GP. The Adam optimizer is used for training. Learning rates of 5e-3 and 1e-2 are used for the training of the deep mean and deep kernel, respectively. These values are hand-picked by inspecting the training loss and are not carefully tuned. For training the mean, 10$\%$ of the training data is used for validation and early stopping based on the validation loss with patience of 5 is used to select the training epochs. Early stopping with a patience of 5 based on the training loss is used to select the training epochs for the deep kernel. We also apply data augmentation, where for both mean and kernel training, random vertical flips of the images are used since flipping vertically would not change the predicted volume. For mean training, random hue jitter and random depth noise are also applied. The training process takes less than 30 minutes on a hardware setup consisting of an i7-9800x CPU, a 2080Ti GPU, and 64GB of RAM.

\subsection{Simulated Experiments}
We perform 2 types of simulated experiments, \textit{prediction accuracy} and \textit{simulated deployment}, on a static test database to evaluate the performance of our methods against the state-of-the-art. The test database consists of 100 randomly chosen scoops on each of the 16 testing terrains

For prediction accuracy, we are evaluating how well each model predicts scoop volume in the $k$-shot setting. 80 samples from each terrain's data are first randomly drawn to form the query set. The support set is randomly drawn from the remaining 20 samples. The prediction accuracy in terms of mean absolute error (MAE) of the model on the query set given the support set is evaluated. 

For simulated deployment, we evaluate how the model's prediction accuracy impacts adaptive decision-making performance. In this experiment, we implement a policy that only selects from the 100 actions in the dataset for the given test terrain, and the robot receives the corresponding reward observed in the dataset. A trial begins by observing a single RGB-D image as input, and the agent executes the policy until the sample reward is above a threshold $B$.  $B$ is customized for a given terrain and is defined as the 5th largest reward in that terrain's dataset. The Single Cardboard Sheet terrain is excluded in these experiments because it is not scoopable.

\subsubsection{Baselines}
We compare our method against two state-of-the-art meta-learning methods and one non-adaptive baseline. The first method is DKMT~\cite{Fortuin2019}, with the same deep network architecture. A learning rate of 1e-2 and early stopping with patience of 5 based on the training loss is used to select the epochs. The second method is conditional neural processes (CNP)~\cite{garnelo2018CNP}, which is a non-kernel-based approach that learns a task representation using the support set and conditions the prediction on the query set on the learned task representation. Neural networks of similar width and depth compared to our proposed neural network architecture are used. During each epoch in meta-training, each task is randomly split into a support set of 5 samples and a query set of 95 samples. The training loss is the NLML on the query set on each task given the support set. Early stopping with a patience of 5 based on the validation loss on 6 randomly sampled validation tasks is used to select the training epoch. For the 0-shot case in CNP, a zero task representation is used. The last baseline is a non-adaptive baseline that only uses the deep mean function of our model.

\subsubsection{Results}
\begin{figure}[]
\centering
    \includegraphics[trim=0cm 0cm 0cm 1cm,clip,width=0.9\linewidth]{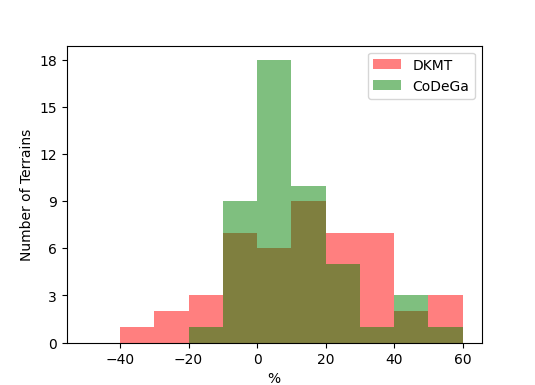}
    \caption{Distribution of MAE reduction from 0-shot to 10-shot support sets (10\% indicates that the 10-shot MAE is 90\% of the 0-shot MAE), over all testing terrains, and aggregated over 3 trained models.}
    \label{fig:MAE_distribution}
\end{figure}

The results are summarized in Tab.~\ref{tab:simulation_accuracy} and Tab.~\ref{tab:simulation_rollout}. Each model is trained 3 times with different random seeds and average results across all tasks aggregated over 3 random seeds are reported. For prediction accuracy, in addition to the average MAE on all query data, we also report the average MAE of the 5 largest samples in the query set of each terrain because it is important to predict well on the ``good samples'' in a reward-maximizing decision-making setting. To compare DKMT and CoDeGa on a finer scale, we plot the distribution of MAE percentage reduction of each terrain from 0-shot to 10-shot support sets on all test terrains, aggregated over all 3 trained models, in Fig.~\ref{fig:MAE_distribution}. For simulated deployment, the adaptive methods are used by a UCB decision maker with $\gamma = 2$, while the non-adaptive baseline sorts actions by the reward predicted by the mean model and greedily proceeds down the list. 
\begin{table}[]
    \renewcommand{\arraystretch}{1.2}
    \centering
    \footnotesize
    \setlength{\tabcolsep}{4pt}
    \begin{tabular}{@{}lcccccc@{}}
        \toprule
         % & \multicolumn{6}{c}{\bf MAE }\\
        % \cmidrule(lr){2-7}
        {\bf Method} & {\bf 0-shot} &  {\bf 5-shot} & {\bf 10-shot} & {\bf 0-shot$^*$} &  {\bf 5-shot$^*$} & {\bf 10-shot$^*$}\\
        \midrule
        CoDeGa  & 27.4 & 24.7 & 23.8  &68.4& 61.3 & 60.8\\
        DKMT~\cite{Fortuin2019} & 25.8 & 22.1 & 21.3 &103.4& 83.6 & 80.1\\
        CNP~\cite{garnelo2018CNP}& 25.7 &25.1 &25.0 &101.4&100.4& 100.5 \\
        Non-adaptive & 27.4 & 27.4 & 27.4 & 68.4& 68.4& 68.4\\
        
        \midrule
        %\bottomrule
    \end{tabular}
    \caption{Average Prediction MAE. Averages taken across all test terrains and across 3 random seeds. $^*$ indicates MAE over the 5 samples of largest volume in the query set, averaged across all test terrains.}
    \label{tab:simulation_accuracy}
\end{table}

From Tab.~\ref{tab:simulation_accuracy}, we observe that DKMT and CoDeGa have similar MAE reduction from 0-shot to 10-shot on average, with DKMT slightly outperforming it. However, CoDeGa outperforms DKMT in terms of prediction accuracy on the good samples. In addition, from Fig.~\ref{fig:MAE_distribution}, we find that DKMT exhibits a high variance, even \textit{degrading} significantly in performance for some terrains from 0-shot to 10-shot adaptation. CoDeGA outperforms DKMT on the simulated deployment experiments. On the Single Rocks testing terrain where DKMT suffers the largest degradation, BO with the DKMT model takes as many as 44 attempts to reach the threshold in for one of the random seeds. This is due to incorrect correlations between low-quality support set samples and samples that are potentially of high quality. Both CoDeGa and DKMT outperform CNP and the non-adaptive approach significantly on both prediction accuracy and simulated deployment. 

\begin{table}[]
    \renewcommand{\arraystretch}{1.2}
    \centering
    \footnotesize
    \begin{tabular}{@{}lcc@{}}
        \toprule
        {\bf Method} & {\bf Avg. Attempts}  & {\bf Max. Attempts} \\
        \midrule
        CoDeGa  & 5.2 & 28\\
        DKMT~\cite{Fortuin2019} & 6.9  & 50\\
        CNP~\cite{garnelo2018CNP}& 9.6  & 40 \\
        Non-adaptive & 8.3 &  57 \\
        
        \midrule
        %\bottomrule
    \end{tabular}
    \caption{Simulated deployment results. Average and Max are taken across all testing terrains excluding the Single Cardboard Sheet terrain, and across 3 random seeds. }
    \label{tab:simulation_rollout}
\end{table}

\subsection{Physical experiments}
Lastly, we evaluate the real-world performance of our method in physical deployments. Here, the robot has a larger action set, it executes the scooping sequence as determined by the optimizer, and each action introduces terrain shifting for the subsequent action, so the RGB-D image is re-captured after every scoop. Policies are deployed on the same 15 testing terrains as the simulated experiments and with the same termination threshold $B$. For each trial a budget of 20 attempts is enforced, beyond which the trial is considered a failure.

The action set is a uniform grid over the action parameters, with 15 $x$ positions (3\,cm grid size), 12 $y$ positions (2\,cm grid size), 8 yaw angles, 4 scooping depths, and 2 stiffness, totaling 11520 actions. If robot trajectory planning fails for a scooping action, the next action that has the highest score is selected until planning succeeds.

We compare our proposed method (Ours) to the Non-Adaptive baseline, i.e. only the deep mean, and a volume-maximizing (Vol-Max) policy, where the action is chosen to maximize the intersection between the scoop's swept volume and the terrain following a strategy proposed recently in the excavation literature~\cite{yang2021optimization}. We note that Vol-Max also does not adapt. Ours uses a UCB decision maker with $\gamma = 2$ and the  CoDeGa model, while Vol-Max and Non-Adaptive use a greedy decision maker. 

Each method is run on each terrain three times. Ours and Non-Adaptive are tested with three models trained with different random seeds, while Vol-Max is simply tested 3 times. When deploying the policies on a testing terrain, the terrain is manually reset at the start of each deployment so that surface features are consistent across trials. Note that slight terrain variations are introduced naturally during the reset. 

\begin{table}[]
    \renewcommand{\arraystretch}{1.2}
    \centering
    \footnotesize
    \begin{tabular}{@{}lccc@{}}
        \toprule
        {\bf Method}  & {\bf Avg. Attempts}  & {\bf Max. Attempts} & {\bf Success Rate}  \\
        \midrule
        Ours & 3.1& 16 & 100\% \\
        Vol-Max & 7.3& 20 &91.1\% \\
        Non-adaptive & 6.2& 20 &84.4\%\\
        \midrule
        %\bottomrule
    \end{tabular}
    %\vspace{0.1cm}
    \caption{Results from physical experiments over 3 trials per method on 15 testing terrains. The robot is limited to 20 attempts. }
    \label{tab:actual_deployment}
\end{table}

\begin{figure}[h!]
\centering
    \includegraphics[trim=2cm 15cm 0.5cm 0.7cm,clip,width=\linewidth]{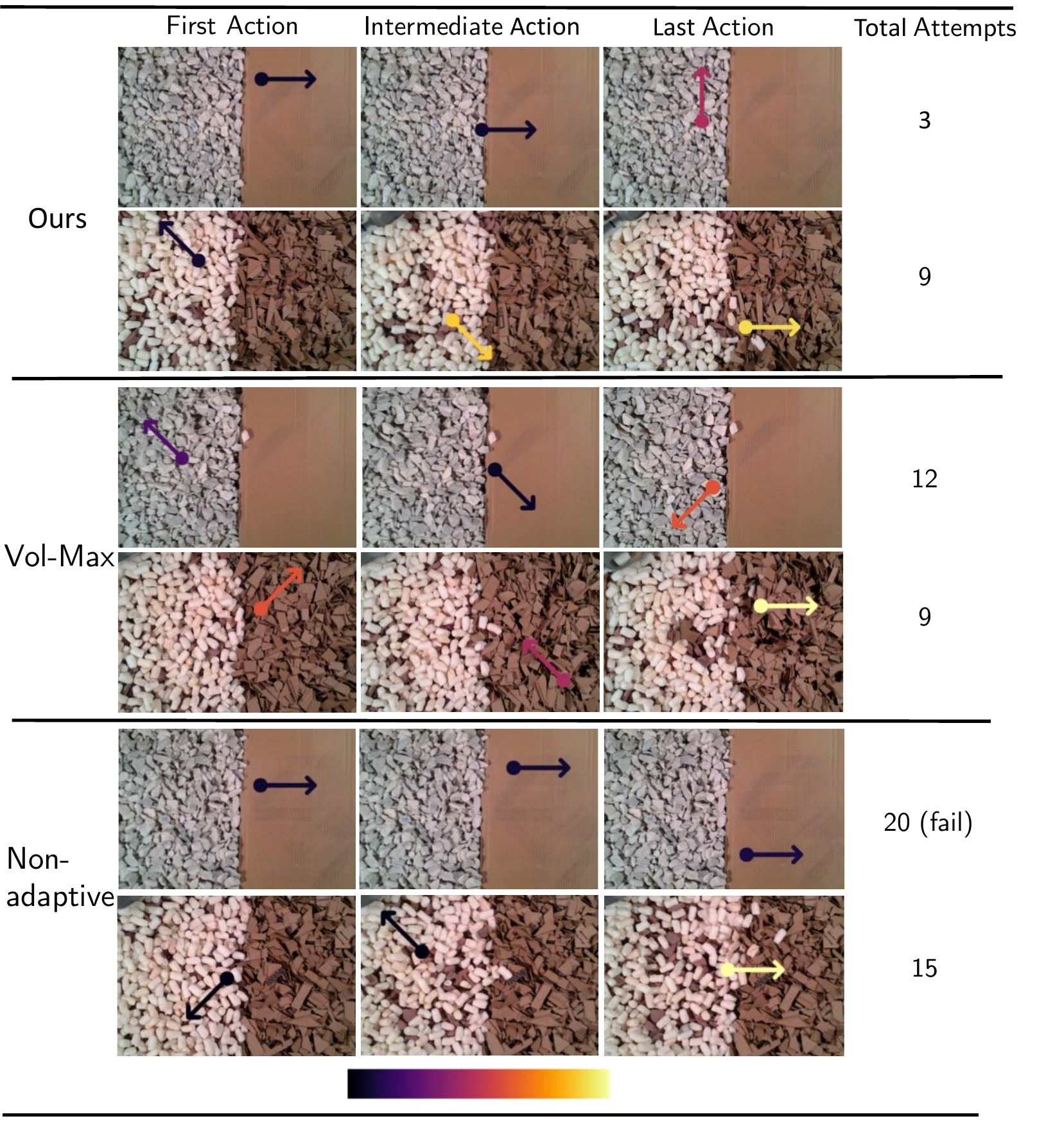} \put(-100,13){\scalebox{0.8}{90\,cm$^3$}} \put(-190,13){\scalebox{0.8}{0\,cm$^3$}}
    \caption{Example physical trials on two terrains comparing our method, the Non-Adaptive baseline, and Vol-Max. The terrain on the top is Partition with Gravel and Cardboard Sheet, while that on the bottom is Layers with Packing Peanuts over Slates on the left and Shredded Cardboard on the right. The arrow indicates executed action while the color indicates the scooped volume.}
    \label{fig:physical_rollout}
\end{figure}

The average and max number of attempts before termination and success rates for all methods are reported in Tab.~\ref{tab:actual_deployment}. Our method outperforms the other two baselines significantly, achieving a 100\% success rate. 
We show two representative trials for each of the three methods in Fig.~\ref{fig:physical_rollout}. On the Partition with Gravel and Cardboard Sheet terrain, the deep mean function predicts higher volumes on the Cardboard Sheet, but our method is able to quickly adapt.  Vol-Max also succeeds, but requires more attempts since maximizing volume is not the most optimal policy to scoop Gravel.  Non-adaptive causes the decision-maker to repeatedly select scoops on the Cardboard Sheet and eventually fail.  On the second terrain, Layers with Packing Peanuts over Slates and Shredded Cardboard, the deep mean function predicts higher volumes on Packing Peanuts, but due to the layer of Slates underneath the scoop jams easily. Our method adapts to this observation in a few attempts.  Vol-Max is able to perform similarly well as our method because Shredded Cardboard has more prominent terrain features, resulting in large intersection volumes. As a result, it always selects to scoop on Shredded Cardboard but takes 9 attempts to obtain high volumes because Vol-Max ignores the arrangement of granular particles, which has a big effect on the scoop outcome. Non-Adaptive takes many samples on Packing Peanuts, but eventually stops because the Slates become exposed, and Slates are in the training database and predicted to yield low volume.

Although our method achieves a 100\% success rate, it is possible to have arbitrary materials, e.g. material with rainbow colors, where the kernels correlate support data samples poorly. In such cases, it is desirable to adapt the kernels online and we leave this to future work.

\section{Conclusion and Future Work} 
\label{sec:conclusion}
This paper introduced a novel method for granular material scooping under domain shift that uses a vision-based deep GP method and a Bayesian optimizer to adapt quickly to small amounts of on-line data. Our novel meta-training procedure, Deep Meta-Learning with Controlled Deployment Gaps, simulates deployment gaps to better train a deep kernel to cope with large domain gaps than the state-of-the-art. We demonstrate in real-world experiments that our proposed approach quickly achieves large scoop volumes on terrains that are drastically different from those seen in training, and significantly outperforms non-adaptive methods. 

In the future, we would like to explore more complex rewards, such as the outcome of a scientific assay from a sample analysis instrument. In addition, the force and torque the scoop experiences when executing a scoop action could be very informative about the underlying terrain and we would like to condition the scooping policy on these as well. Finally, our current method has no control over the controller that tracks the scooping trajectory other than the stiffness parameter. We would like to investigate feedback controllers that can react and adapt to terrain in real-time during scooping to further facilitate adaptation.  

\section*{Acknowledgment}
This work is supported by NASA Grant 80NSSC21K1030.

\bibliographystyle{plainnat}
\bibliography{references}

\end{document}